\documentclass[letterpaper, 10 pt, journal, twoside]{IEEEtran}
\IEEEoverridecommandlockouts
\usepackage{cite}
\usepackage{amsmath,amssymb,amsfonts}
\usepackage[ruled]{algorithm2e}
\usepackage{algorithmic}
\usepackage{graphicx}
\usepackage{textcomp}
\usepackage{xcolor}
\usepackage{tikz}
\usetikzlibrary{matrix, calc}
\usepackage{caption}
\def\BibTeX{{\rm B\kern-.05em{\sc i\kern-.025em b}\kern-.08em
    T\kern-.1667em\lower.7ex\hbox{E}\kern-.125emX}}
    
\usepackage[compact]{titlesec}    
\begin{document}

\setlength{\parskip}{0cm}
\setlength{\parindent}{1em}
\titlespacing{\section}{0pt}{1.75mm}{0.0mm}
\titlespacing{\subsection}{0pt}{1.75mm}{0.0mm}

\title{Learning to Assist Drone Landings\\
}
\author{Kal~Backman,~Dana~Kulić~and~Hoam~Chung
\thanks{Manuscript received: October, 15, 2020; Revised January, 17, 2021; Accepted February, 11, 2021.}
\thanks{This paper was recommended for publication by Editor J. Ryu upon evaluation of the Associate Editor and Reviewers' comments.}
\thanks{Authors are with the Department of Electrical and Computer Systems Engineering and the
Department of Mechanical and Aerospace Engineering, Monash University. [Kal.Backman1, Dana.Kulic, Hoam.Chung]@monash.edu.}
\thanks{Digital Object Identifier (DOI): 10.1109/LRA.2021.3062572}
}

\markboth{IEEE Robotics and Automation Letters. Preprint Version. Accepted February, 2021}
{Backman \MakeLowercase{\textit{et al.}}: Learning to Assist Drone Landings} 

\maketitle

\begin{abstract}
Unmanned aerial vehicles (UAVs) are often used for navigating dangerous terrains, however they are difficult to pilot.  Due to complex input-output mapping schemes, limited perception, the complex system dynamics and the need to maintain a safe operation distance, novice pilots experience difficulties in performing safe landings in obstacle filled environments. In this work we propose a shared autonomy approach that assists novice pilots to perform safe landings on one of several elevated platforms at a proficiency equal to or greater than experienced pilots. Our approach consists of two modules, a perceptual module and a policy module.  The perceptual module compresses high dimensionality RGB-D images into a latent vector trained with a cross-modal variational auto-encoder. The policy module provides assistive control inputs trained with the reinforcement algorithm TD3. We conduct a user study (n=33) where participants land a simulated drone with and without the use of the assistant. Despite the goal platform not being known to the assistant, participants of all skill levels were able to outperform experienced participants while assisted in the task.
\end{abstract}


\section{Introduction}
\IEEEPARstart{U}{nmanned} aerial vehicles (UAVs) have become increasingly popular for reconnaissance and inspection tasks, due to their high mobility in traversing dangerous terrains such as mine shafts and tunnels. In such environments, remotely operated UAVs are preferred over full automation, as it is difficult to fully automate human decision making in the challenging and safety-critical context.  However, the high mobility comes at a cost of increased teleoperation complexity. When coupled with poor depth perception from maintaining a safe operation distance and increased disturbances from the ground effect \cite{PaperI1}, performing a safe landing in obstacle-filled environments is difficult for novice pilots. To reduce the cost of training pilots and increase the accessibility of UAV piloting, we seek to find a method that allows novice pilots to safely land at a performance level equal to or greater than expert pilots.

Previous autonomous landing methods \cite{PaperI2, PaperI3, PaperI5} reduce the difficulty and cognitive workload of the task, however these methods rely on knowing the pilot’s goal prior to landing, making them unsuitable for unstructured environments. Such methods also neglect any form of adaptation to the pilot's input, resulting in a loss of control of the UAV and the inability for the pilot to react to external disturbances not perceived by the autonomous system. Shared autonomy systems address these concerns by combining user inputs with that of artificial intelligence to collaboratively complete tasks in an assistive manner. However assistive approaches are limited by the proficiency of the pilot and their ability to timely react under pressure. Further issues arise when actions taken by the assistant and pilot conflict, leading to unsatisfactory flight behaviour and a greater risk to crash.

There are two main challenges when implementing a shared autonomy system: (i) predicting the user’s goal and (ii) deciding how to assist the user based on this prediction. In the context of safely landing a drone, multiple safe landing zones may exist  however, depending on the current and future goals of the pilot, only a subset of these may be suitable. This goal uncertainty  is caused by private information from the pilot being unknown to the assistant.  Although the pilot could explicitly state these goals through alternate interfaces, prior work suggests that explicit communication leads to ineffective collaboration \cite{PaperI7, PaperI8, PaperI9} compared to implicit means. 
 
Once the goal is inferred, the assistant must then decide what degree/type of assistance to provide. Providing insufficient assistance can result in  task failure whilst excessive exerted control, albeit following an optimal strategy, leads users to perceive the system as untrustworthy \cite{PaperI10}. 

The challenges of predicting the pilot’s goal and how much control should be exerted are related to the perspective of the assistant, which views the pilot as providing an approximation of the optimal policy that needs to be fine-tuned. From the perspective of the pilot, two challenges exist: (i) the uncertainty in estimating the state of the UAV and (ii) developing a dynamic model that maps control inputs to physical changes of the UAV’s state.  For the pilot to successfully land the UAV, they must estimate the relative position of the UAV to that of the landing target. Estimation of the UAV’s depth is challenging for pilots, as it is inferred from the perceived size of the UAV, whereas estimation of the UAV’s latitude and elevation are based on the projected location within the pilot’s view plane. 

This leads to poor depth estimation of the UAV and the resultant actions taken by the pilot being misaligned with the intended goal. To successfully operate the UAV, pilots must develop an internal dynamic model that maps control inputs to state transitions. This mapping and dynamic model is developed through accumulated interactions with the environment and is a result of the pilot’s experience. 

\subsection{Problem statement and model overview}
Our goal is to develop an assistive system that allows novice pilots to land a UAV on one of several platforms at a proficiency greater than or equal to that of an experienced pilot without the need to explicitly provide the pilot's intent. We define an experienced pilot as one that can accurately estimate the position of the UAV and reliably land on the centre of a given platform. To control the drone the pilot provides target linear XYZ velocities for the on-board flight controller to follow. The pilot is aware of the desired landing platform but due to maintaining a safe distance, has difficulty in perceiving the depth of the UAV. On the other hand, the assistant understands the control inputs required for a successful landing however it is unaware of the pilot’s goal which must be inferred from observations of the pilot’s commands to the UAV during interaction with the environment. 

\begin{figure}
\centering
\includegraphics[width=1.0\columnwidth]{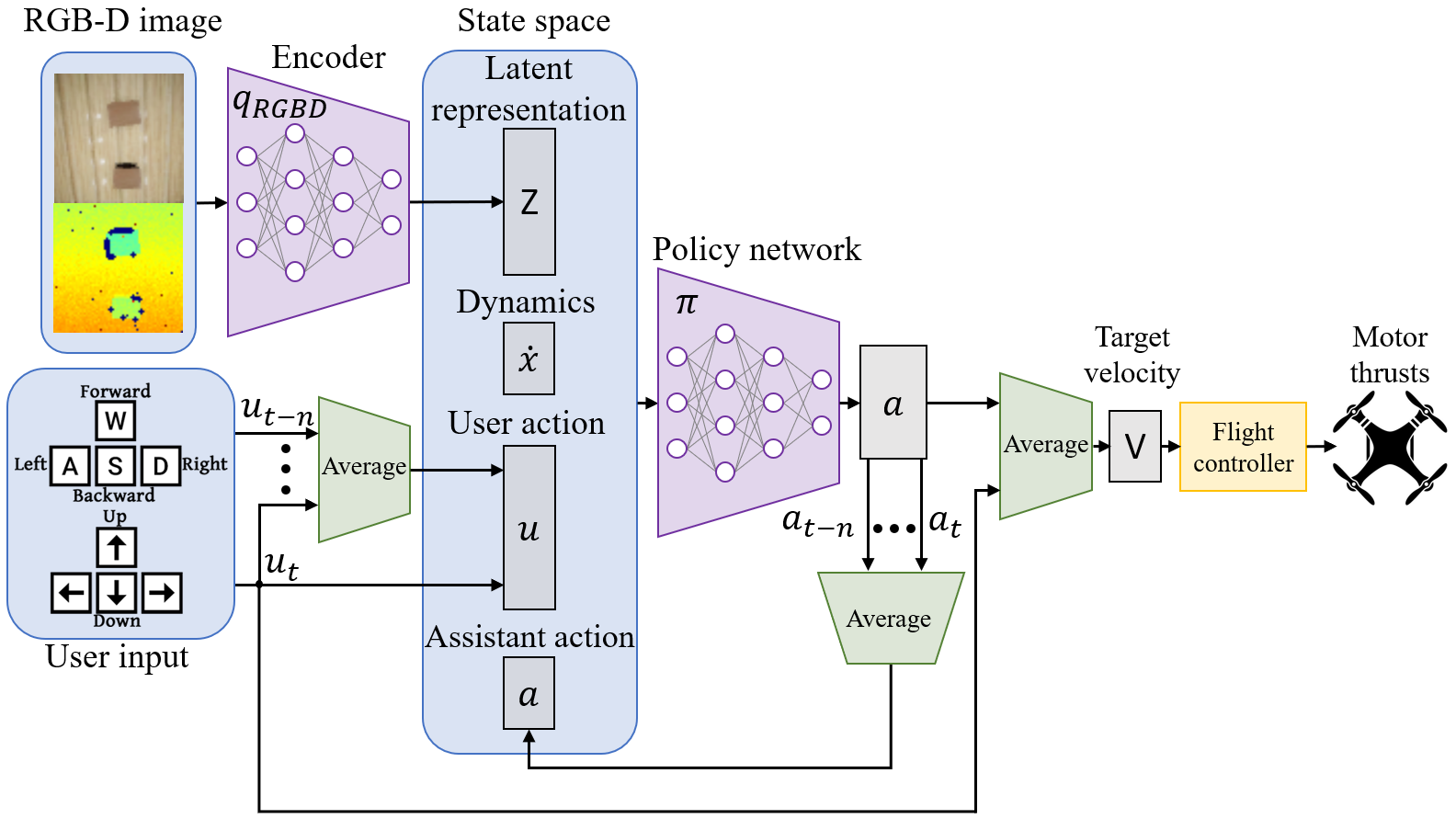}
\caption{System architecture used to assist pilots in landing. Encoder \(q_{RGBD}\) takes an RGB-D image as input and creates a compressed latent representation. The latent representation is then fed into the policy network along with other state variables from which an action is generated. The resultant action is then averaged with the pilot’s action to create a target velocity for the flight controller.}
\label{NetworkSummary}
\end{figure}

Our approach is summarised in Fig.~\ref{NetworkSummary} and is comprised of two components: (i) learning a compressed latent representation of the environment to perceive the location and structure of the landing platforms and (ii) learning a policy network to provide control inputs to assist the pilot in successful landing. The first component takes noisy RGB-D data from a downwards facing camera attached to the UAV and encodes information of the scene and landing platform location into a low-dimensional latent vector representation. The second component takes the latent vector, UAV dynamics, pilot input and the network's previous average action as input to output a target velocity for the UAV. The UAV’s final target velocity is determined by averaging the action of the pilot and that of the network which is then fed into the flight controller.

\subsection{Related work}
Common fully autonomous landing methods use visual servoing \cite{PaperRW1} to approach a predefined target. The limitations of such an approach is that the target must remain in the camera's field of view throughout the task. Model-based approaches are favoured for moving platform tasks \cite{PaperI3, PaperI5}, alleviating the need for continual visual observations by predicting the motion of the UAV and target. Both approaches require that a single unambiguous landing zone exists and is of a predefined shape, making them unsuitable for unstructured environments. Our approach aims to address these concerns by training over a wide variety of landing zones where multiple potential landing zones exist within a single environment.

Learning based approaches \cite{PaperI2, PaperI4} use deep neural networks to learn the necessary control inputs required for successful landing. \cite{PaperI2} uses a hierarchy of deep Q-learning networks (DQNs) to learn an end-to-end mapping from input RGB images to output action velocities. The network was able to learn to detect and land on observed targets however due to the large state-space from end-to-end mapping, the policy required two separate DQNs to manage the detection and landing phase independently. By learning from a latent representation in our proposed approach, a single policy network is required due to the increased sample efficiency of the learning process.

The above autonomous landing approaches lack the ability to handle multiple potential landing zones and are not adaptive to pilot inputs. Previous shared autonomy systems \cite{PaperI13, Paper8, PaperI14} facilitate user adaptability and handle unknown goals but require either a known dynamic model of the world, a set of known potential goals or a set of predefined user policies. These constraints limit the practicality of implementing such approaches due to the difficulty in developing accurate dynamic models and the requirement to pre-define the position of potential landing sites beforehand. Reddy et al. \cite{PaperI15} use shared autonomy to demonstrate a model-free approach using deep Q-learning to assist pilots landing in the lunar lander game. The policy network is unaware of the pilot’s goal, which is inferred from observations of the environment and the pilot’s actions. The drawback of deep Q-learning is that the agent has access to a discrete set of actions which hinders the ability to provide fine control inputs required for safe landing. Instead we implement a policy-gradient reinforcement learning approach to facilitate continuous action space control. 

Reddy et al. \cite{PaperI15} deploy their DQN to a physical UAV landing user study, however the number of participants (n=4) is limited and the landing pad location is included in the state space for the DQN. Sa et al. \cite{PaperCorke} implement shared autonomy for UAV pole inspection tasks, using a reduction in task space controls to constrain the motion of the UAV. The task is made easier for novice pilots by reducing the controllable degrees of freedom so that the UAV is forced to move concentric to the pole whilst automatically controlling the yaw so that the camera is directly facing the inspected pole. 

This paper makes the following contributions: (1) we integrate state-of-the-art learning approaches to infer implicit pilot intent for assisted UAV landing; (2) We provide an extensive validation in a  user study, which demonstrates landing without prior knowledge of the environment or the pilot’s goal in simulation.  Further, the study findings demonstrate that both experts and novices are successfully assisted, and that with assistance the performance of novices approaches that of experts. (3) We provide an initial physical demonstration of the system.

\section{Learning latent space representation}
The purpose of latent space learning is to reduce the time required for the policy network to reach convergence. As the policy network gathers observations in real time, it is costly to explore large state spaces, therefore a reduction in the dimensionality of the state space leads to accelerated convergence \cite{Paper1}. Desired characteristics of a dimensionality reduction technique is that it is smooth, continuous and consistent \cite{Paper2}, so that a policy network can efficiently learn how to perceive the encoded environment. 

\cite{Paper2} introduces a cross-modal variational auto-encoder (CM-VAE) which trains a latent vector representation from multiple data sources. Given a data sample \(x_k\) where \(k\) represents the modality such as a RGB image or an arbitrary sensor reading, the encoder \(q_k\) transforms \(x_k\) into a vector of means \(\mu\) and variances \(\sigma^2\) of a normal distribution, from which the vector \(z\) is sampled,  i.e. \(z \sim \mathcal{N}(\mu, \sigma^2)\). The decoder \(p_l\) takes the latent sample \(z\) and reconstructs to the desired modality \(l\) which is denoted as \(y_l\). As the encoder \(q_k\) maps the input onto a distribution, a regularizing term is required to ensure that the properties of the desired distribution are met. The Kullback-Leibler divergence is used so that the distribution \(q_k\) maps to a normal distribution of \(\mathcal{N}(0,1).\) Traditional VAEs achieve encoding onto a distribution by maximising the variational lower bound on the log-likelihood of the data as seen in \cite{Paper7}. The loss is re-derived in \cite{Paper2} for the CM-VAE to take into consideration different modalities.

\subsection{CM-VAE implementation}
Our implementation of CM-VAE architecture is based on \cite{Paper3}, which uses a CM-VAE to reconstruct RGB images of gates and their pose parameters from a UAV. Our implementation contains a single input modality from a noisy RGB-D camera, denoted as \(x_{RGBD}=[x_{RGB},x_D]\), and two output modalities in the form of a depth map and relative pose of the closest visible landing pad, denoted as \(\hat{y}_D\) and \(\hat{y}_{XYZ}=[\hat{X},\hat{Y},\hat{Z}]\) respectively. Similar to \cite{Paper3}, we use DroNet \cite{Paper4} for the encoder \(q_{RGBD}\), which is equivalent to an 8-layer ResNet \cite{Paper5}. Decoder \(p_D\) takes the full latent space vector \(z\) i.e. \(\hat{y}_D=p_D (z)\), whereas reconstructing \(y_{XYZ}\) we use 
three independent decoders \(p_X\), \(p_Y\) and \(p_Z\), each taking a single element of the first three elements of the latent space vector \(z\) i.e. \(\hat{X}=p_X (z_{[0]} )\), \(\hat{Y}=p_Y (z_{[1]} )\) and \(\hat{Z}=p_Z (z_{[2]} )\).

To generate data to train the network we use the AirSim \cite{Paper6} Unreal engine plugin to capture RGB-D images of generated scenes. Each generated scene consists of five landing platforms that are aligned along the longitudinal direction with random spacing between them. The platform's width, length and height are generated from a uniform distribution and then each platform is independently shifted and rotated about the yaw axis. Textures are then applied to the platforms, walls and floor from a total of 52 materials. The camera is swept through the scene where 50 RGB-D images and relative landing platform locations are recorded before a new scene is generated. 

Depth images provided by AirSim are free of noise which is not indicative of real-life sensors. Therefore, a noise generating function \(G\) is used to create training data, \(x_D=G(y_D)\). Our implementation differs from \cite{Paper3} as the objective of our CM-VAE is to not identically reconstruct the input but to reconstruct a denoised version of the input, making our model a denoising CM-VAE.

To train the network, we follow Algorithm 1 outlined in \cite{Paper2} where we have three losses: (i) MSE loss between the true depth map and reconstructed depth map \((y_D  ,\hat{y}_D\)), (ii) MSE loss between the true relative landing platform parameters and the estimated parameters  \((y_{XYZ}  ,\hat{y}_{XYZ})\) and (iii) Kullback-Leibler divergence loss. The total loss is the sum of the three weighted MSE losses where the weights for \((y_D,\hat{y}_D)\),  \((y_{XYZ},\hat{y}_{XYZ})\) and KL loss are 1, 2 and 4 respectively. For every input sample \(x_{RGBD}\), there is a corresponding true output sample for each of the modalities, therefore for each training iteration the weights of \(q_{RGBD}\), \(p_D\), \(p_X\), \(p_Y\) and \(p_Z\) are updated. Results of the training process will be presented in Section \ref{Result_CMVAE}.

\section{Policy learning - TD3}
The aim of the policy network is to assist the pilot in landing on the desired landing platform by outputting a target velocity for the UAV that is combined with the user’s current input. Following \cite{Paper8} formalizes shared autonomy systems as a partially observable Markov decision process (POMDP), where the user’s goal is a hidden state that must be inferred by the agent. We define our problem as a POMDP where the set of all possible states is denoted as \(\mathcal{S}\). The pilot’s action  is part of the current state \(s\). The set of actions the agent can take is denoted as \(\mathcal{A}\). We define the state transition process as a stochastic process, due to the uncertainty in predicting UAV dynamics from turbulence caused by the ground effect, as \(T : \mathcal{S} \times \mathcal{A} \times \mathcal{S} \rightarrow [0, 1]\). The reward function is defined as \(R : \mathcal{S} \times \mathcal{A} \times \mathcal{S} \rightarrow \mathbb{R}\), the set of observations as \(\Omega\) with the observation function \(\mathcal{O} : \mathcal{S} \times \Omega \rightarrow [0, 1]\) and discount factor \(\gamma \in [0,1]\). Given an optimal policy \(\pi : \mathcal{S} \times \mathcal{A}\) that maximises the expected future discounted reward of the Bellman equation

\setlength\abovedisplayskip{1.15mm}
\setlength\belowdisplayskip{1.15mm}
\begin{equation}
Q^\pi(s,a)=R(s,a,s') + \gamma E[Q^\pi(s',a')], \label{BellmanEq}
\end{equation}

the goal of reinforcement learning is to find the closest approximation of the optimal policy.

To approximate the optimal policy, we implement the reinforcement algorithm TD3 \cite{Paper9} as it allows for a model-free continuous action space control whilst addressing the instability concerns of its predecessor DDPG \cite{Paper10}. TD3 is an off-policy actor-critic method that uses double Q-Learning \cite{Paper11} by taking the minimum value between two critics in estimating the Q-value for the next state action pair. Clipped noise is added to the target actor’s action to smoothen the estimated Q-value around the action given a state. The full algorithm is  described in Algorithm 1 in \cite{Paper9}.

\subsection{TD3 implementation} \label{TD3Implementation}
To train the agent we utilize AirSim as our simulation environment for modelling of UAV dynamics. We generate a scene containing a total of five potential landing platforms aligned longitudinally with random positions within a confined boundary.
As training reinforcement learning algorithms takes substantial time, we developed a simulated user instead of human participants for use in training. The simulated user models several behavioural flight and landing styles including attempting to land on incorrect adjacent landing platforms before correcting itself. For each landing sequence, the simulated user is given an initial estimate of where it believes the true goal \(G\) is, which is denoted by \(\hat{G}\).  We characterise a simulated user with two parameters \(\alpha \in [0,1]\), which describes the simulated user’s conformance to the agent’s actions and \(\beta \in [0,1]\), a measure of the user’s skill which is the ability to improve its own estimate of the goal. Both \(\alpha\) and \(\beta\) affect the simulated user’s current estimate of the goal position by:
\begin{equation}
\hat{G}_{i+1} = \hat{G}_i + \alpha \frac{a}{K_\alpha} + \beta \frac{G - \hat{G}_i}{K_\beta}, \label{GoalUpdateEq}
\end{equation}
where \(a\) is the action taken by the agent and \(K_\alpha\) and \(K_\beta\) are both scaling constants. 

For each iteration, the agent receives the mean latent space vector from the encoder \(q_{RGBD}\), the UAV dynamics, the simulated user’s current action and the average user and agent actions from the previous 1 second interval. The action taken by the agent is a vector in \(\mathbb{R}^3\) which is averaged with the user’s input \(u\) given that the agent’s action has a magnitude greater than 0.25. We ignore small actions taken by the assistant  to allow the assistant to relinquish control to the pilot. The resultant vector \(V_t\) is the target velocity vector which the UAV should follow.

\begin{equation}
    V_t =
    \begin{cases}
        \frac{u + a}{2}     &\text{if \(|a| > 0.25\)} \\
        u                   &\text{else}
    \end{cases}
\end{equation}

To emulate disturbances from general turbulence and the ground effect, we add normally distributed noise to the UAV’s target velocity where the variance of the noise is dependent on the distance to the ground. By adding noise to the UAV’s target velocity, the simulated user and agent will experience uncertainty in state transitions which better reflects real world control of UAVs. We implement an Ornstein–Uhlenbeck process \cite{PaperOrnstein} for exploratory noise which is added to the action taken by the agent and is decayed after successive landings.

Our reward function utilizes three different rewards: (i) the landing distance of the UAV to the centre of the goal platform, (ii) the difference between the distance to the true goal for the current state \(s_i\) and the previous state \(s_{i+1}\), given that the distance between the true goal and \(s_{i+1}\) is below a threshold and (iii) the magnitude of the action taken by the agent. To take advantage of the off-policy approach we load the experience replay buffer with previously collected state action transitions from previous models. We use a learning rate for the actor and critic of 1e-3, discount factor of 0.99, batch size of 100 and a target actor and critic update rate of 0.01.  Results of the training process will be shown in Section \ref{Result_TD3}.

\section{User study}
To assess the performance of the network for users of varying skill levels, we conducted a user study where participants performed the task in a simulated environment. The user study was conducted remotely in an unsupervised manner where participants accessed the simulated environment through a web browser. Participants were first shown an introductory video that explained the task and study. The task assigned to participants was to land the simulated drone on one of five platforms in the environment.  Participants could move the drone using the W-A-S-D keys for latitude and longitudinal control and the up and down arrows for elevation. 

\begin{figure}[!t]
\centering
\includegraphics[width=1.0\columnwidth]{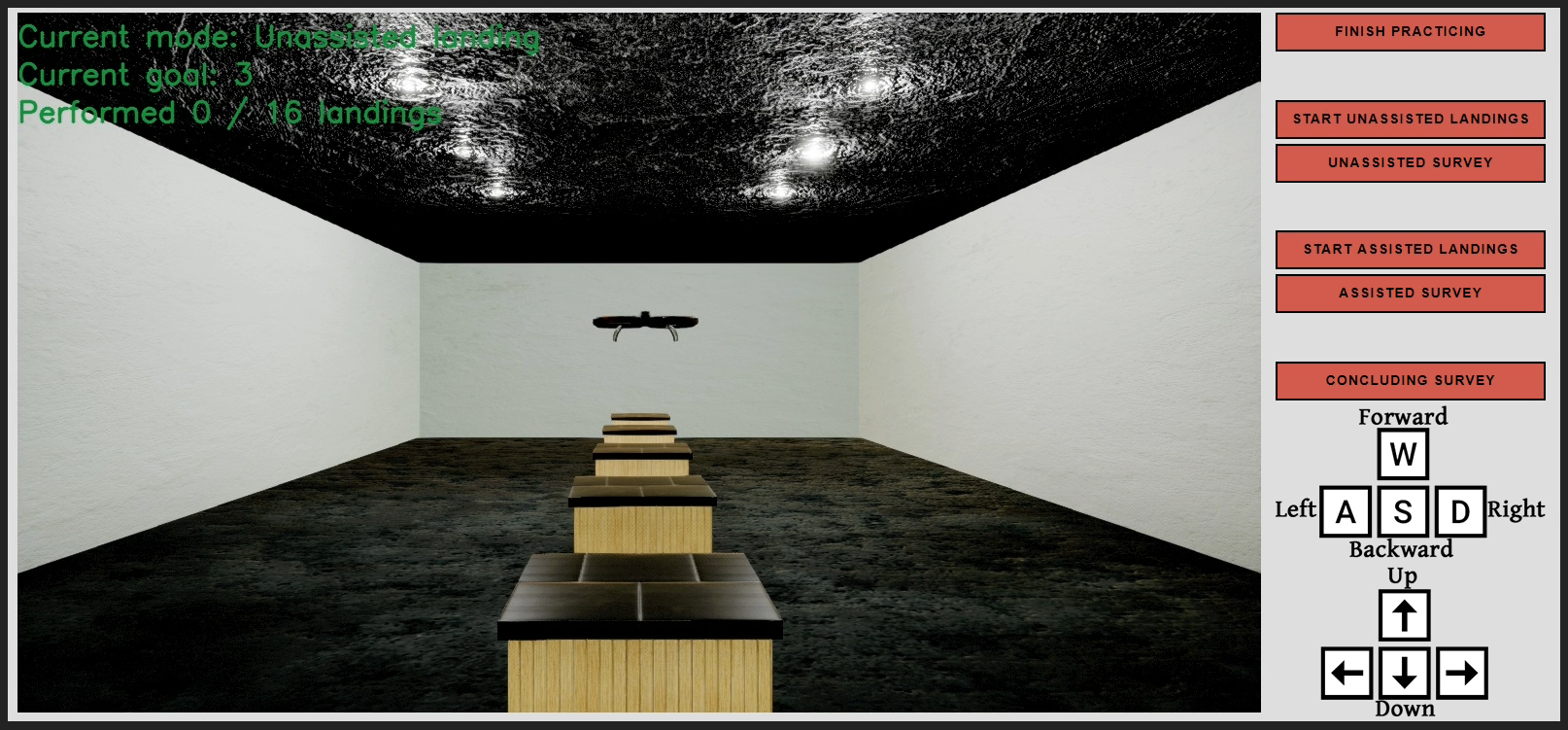}
\caption{Example of a participant performing the study in their web browser.}
\label{UserStudyImage}
\end{figure}

Each participant performed the task in an identical environment, shown in Fig.~\ref{UserStudyImage}. Participants were permitted to perform a maximum of 14 practice landings to understand the requirements of the task and control scheme. During the practice landings the goal platform was randomly assigned, and participants were given feedback after each landing by displaying their relative position to the landing platform. Participants were then assigned to perform the task either unassisted or assisted based on their participant ID number. Each participant was instructed to perform a total of 16 landings in a predefined sequence where no feedback about their performance was given. 

After completing the initial 16 landings, participants were given a system usability scale (SUS) survey \cite{Paper12} to assess the mode they just completed. The task was then repeated using the same predefined landing sequence but for the opposite mode of flying which was again followed by the SUS survey. After completing all tasks the participants were asked to fill out the concluding survey, comprised of twenty-one questions. Seven questions related to the participant's experience in piloting physical drones and video games whilst the other fourteen questions queried the participant's perception of the task. The full list of questions is in Table.~\ref{UserStudyDemoTable} \&~\ref{SurveyTable2}, Sec. \ref{UserStudyResults}. 

\section{Results}
\subsection{CM-VAE results} \label{Result_CMVAE}
\begin{figure}
\centering
\includegraphics[width=1.0\columnwidth]{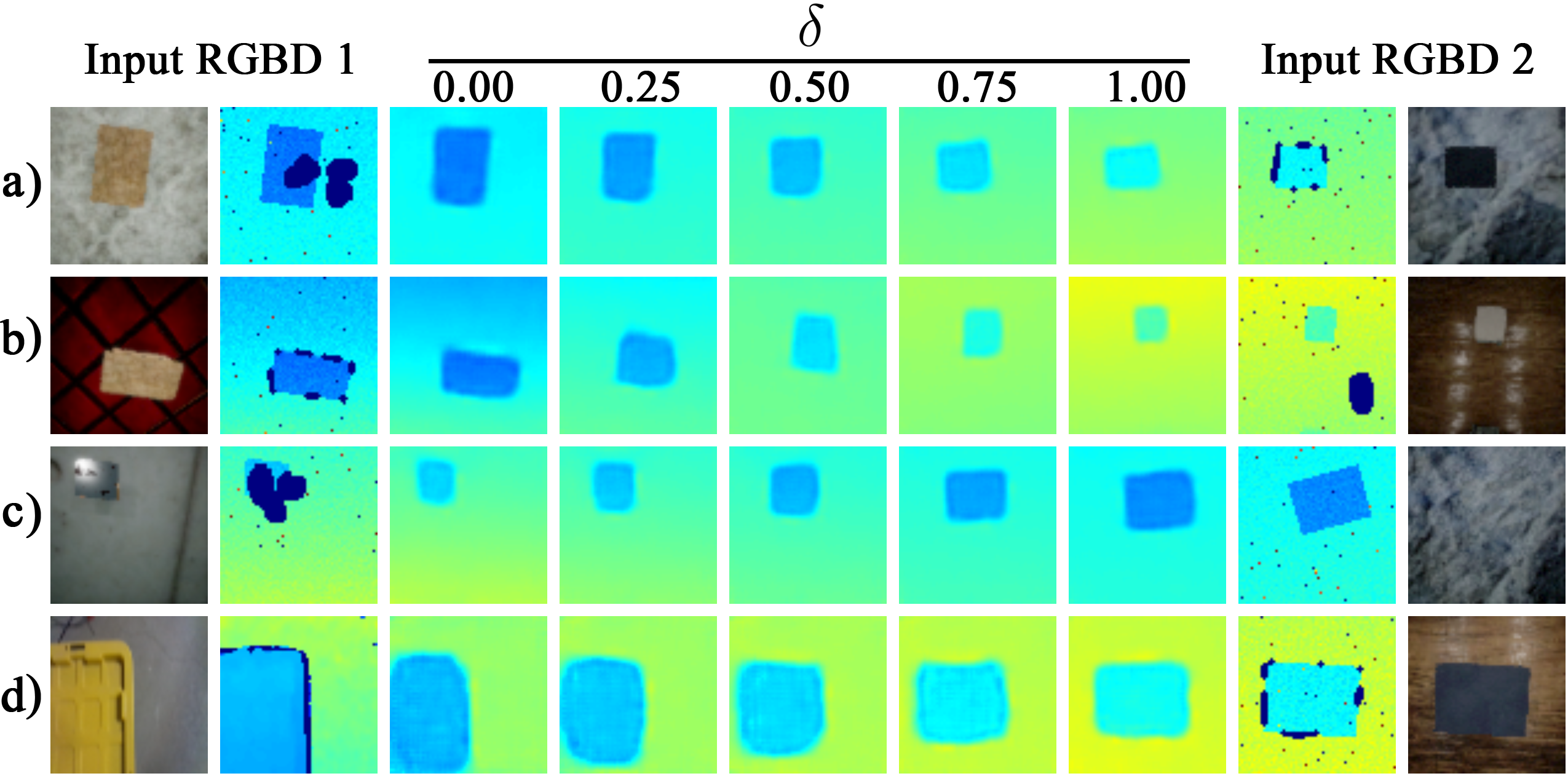}
\caption{Visualisation of latent space reconstruction by interpolating between two reference RGB-D images.}
\label{LatentInterpolation}
\end{figure}

The model was trained with 50k examples of 64\(\times\)64 RGB-D images. A separate dataset is used to demonstrate the interpolation between latent vectors. Fig.~\ref{LatentInterpolation} shows the reconstruction of linearly interpolated latent vectors using the encoder’s mean vector from two separate RGB-D images. The decoder's input vector is determined by \(\mu _i= \mu _1+\delta (\mu _2-\mu _1)\) where \(\mu_1\) and \(\mu_2\) represent the mean vector from the encoder of image 1 and image 2 respectively. Image pairs a) and b) demonstrate the ability to  interpolate between simple geometric transformations due to changes in the landing platform shape and relative position. Image pair c) demonstrates the ability to reconstruct the scene under harsh conditions where input image 1’s depth map is degraded with missing depth values around the landing platform, forcing the encoder to utilize RGB information to recover the depth. Input image 2 in image pair c) shows the alternate case where relying on RGB information is unreliable due to the landing platform and the floor’s texture being identical. Image pair d) shows the ability to transfer to real images where input image 1 is taken from an Intel RealSense D435i camera.

\subsection{TD3 simulation results} \label{Result_TD3}
\begin{figure}[!t]
\centering
\includegraphics[width=1.0\columnwidth]{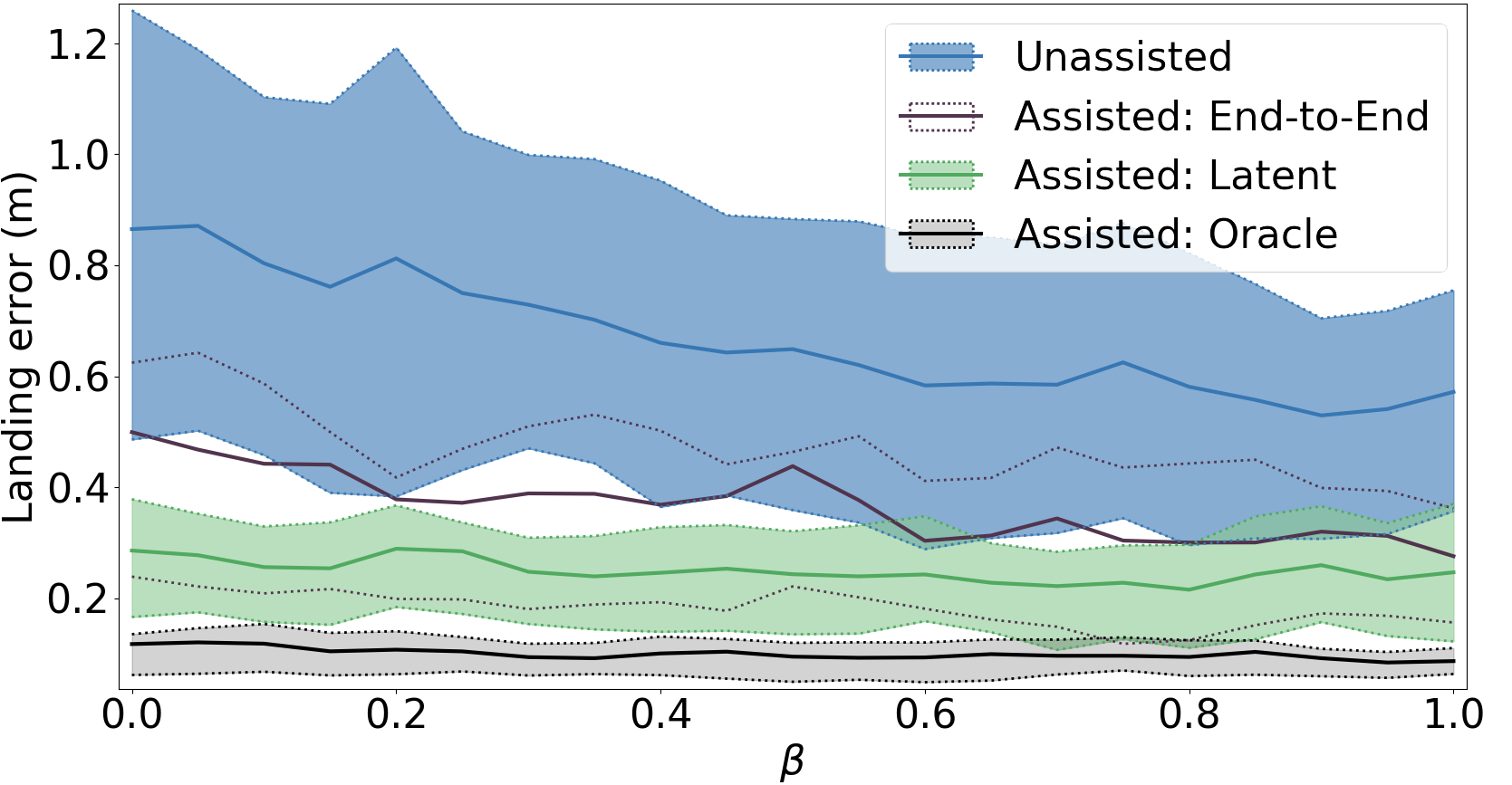}
\caption{Performance of the assistants with the simulated user where \(\beta\) acts as a measure of user proficiency.}
\label{SimulatedUserEval}
\end{figure}

To quantify the level of assistance provided by the assistant, an experiment was run where a simulated user piloted the UAV for incremental \(\beta\) values. The simulated user performed 40 landing sequences for each \(\beta\) value where \(\beta\) was increased in 0.05 increments at an \(\alpha\) value of 0.5. The experiment was first performed without the assistant and then repeated under three assisted conditions. (i) End-to-end: the assistant was trained with the full RGB-D image included in its state space, using an equivalent sized convolutional network as the latent approach. (ii) Latent: the assistant was trained as outlined in Section \ref{TD3Implementation}. (iii) Oracle: the assistant was trained with the true goal location within its state space, providing a baseline to assess the effect of uncertainty about the pilot’s goal on the assistant. The results of the experiment can be seen in Fig.~\ref{SimulatedUserEval}.

Whilst unassisted the average error of the simulated user was 0.67m$\pm$0.40. When assisted with the end-to-end, latent and oracle assistant, the average landing errors were 0.37m$\pm$0.33, 0.25m$\pm$0.15 and 0.10m$\pm$0.07 respectively. The latent assistant enabled simulated users of all skill levels to outperform an unassisted skilled user with greater consistency. As expected, the best performance is achieved by the oracle configuration, indicating the upper bound on the performance improvement that could be achieved if the pilot's intent was perfectly known. The end-to-end assistant was able to provide assistance however performed the worst out of the assisted approaches despite being trained for over twice as many epochs.

\subsection{User study results} \label{UserStudyResults}
The user study was conducted with the approval of Monash University Human Research Ethics Committee (MUHREC), number 25454.  A total of 33 participants completed the user study. The summarised pilot performances can be viewed in Table.~\ref{UserStudyResultTable}. Participants performed the task with greater proficiency whilst assisted, reducing their median error by over a third and increasing their landing success rate by a factor of three. The performance results are split into two categories based on whether the pilot landed on the instructed platform or to the nearest platform. The error for a given landing is represented as the Euclidean distance of the drone’s centroid to the centre of the platform along the latitude and longitudinal axis. A successful landing is recorded when all four legs of the drone contact the landing platform, while a failure is recorded if the drone falls below a threshold height and not all legs are contacting the platform. Of the 528 unassisted and 528 assisted landings, 10 unassisted landings and 16 assisted landings are excluded from statistical analysis due to the participant immediately landing on the starting platform. 
\begin{table}[!b]
\captionof{table}{User study results summary}
\label{UserStudyResultTable}
\begin{tabular}{|p{2.4cm}|c|c|c|c|}
 \hline
 \multicolumn{1}{|c|}{} & \multicolumn{2}{c|}{Specified platform} & \multicolumn{2}{c|}{Nearest platform}\\
 \hline
 \multicolumn{1}{|c|}{} & \multicolumn{1}{|c}{Unassisted} & \multicolumn{1}{c|}{Assisted} & \multicolumn{1}{|c}{Unassisted} & \multicolumn{1}{c|}{Assisted}\\
 \hline
  \multicolumn{1}{|p{2.4cm}|}{Success rate} & \multicolumn{1}{|c}{26.44\%} & \multicolumn{1}{c|}{\textbf{74.85\%}} & \multicolumn{1}{|c}{29.50\%} & \multicolumn{1}{c|}{\textbf{93.76\%}}\\
  \multicolumn{1}{|p{2.4cm}|}{Median error} & \multicolumn{1}{|c}{1.06m} & \multicolumn{1}{c|}{\textbf{0.29m}} & \multicolumn{1}{|c}{0.88m} & \multicolumn{1}{c|}{\textbf{0.25m}}\\
  \multicolumn{1}{|p{2.4cm}|}{Mean error} & \multicolumn{1}{|c}{1.35m} & \multicolumn{1}{c|}{\textbf{1.02m}} & \multicolumn{1}{|c}{0.98m} & \multicolumn{1}{c|}{\textbf{0.28m}}\\
  \multicolumn{1}{|p{2.4cm}|}{Variance} & \multicolumn{1}{|c}{\textbf{1.24m$^2$}} & \multicolumn{1}{c|}{2.33m$^2$} & \multicolumn{1}{|c}{0.46m$^2$} & \multicolumn{1}{c|}{\textbf{0.05m$^2$}}\\
 \hline
\end{tabular}
Bold values represent a statistically significant difference between the two samples in terms of the most favourable value for the given metric at a significance level of \(\alpha\) = 0.01. 
\end{table}

Two-sample statistical tests are performed at a 99\% confidence level to test if a statistically significant difference exists between performing the task unassisted and assisted for a given metric. The test to determine a significant difference for the success rate, median, mean and variance of the error are the McNemar test, Mood’s median test, t-test and F-test respectively. A significant difference was found for all metrics. 

To account for additional effects such as learning and prior experience, a multi-variable regression model is used. The model contains three binary independent variables: (i) whether or not the landing was performed with assistance, (ii) whether or not the landing was performed in the initial half of the study and (iii) whether or not the participant is considered an experienced participant. An experienced participant is defined as one who rated themselves highly in questions 2, 3, 5 and 6 of Table.~\ref{UserStudyDemoTable}. A threshold rating was chosen so that a third of the participants were considered experienced. The binary value associated with human learning (ii) was found to be statistically insignificant at \(p = 0.595\), indicating that insufficient evidence exists to infer that participants improved after performing the initial 16 landings. Participant’s previous experience (iii) was found to be marginally significant (\(p = 0.065\)). A significance value of \(p < 0.001\) was found for (i) suggesting that assistance significantly affected performance whilst controlling for human learning and participant previous experience.   

\begin{figure}[b]
\centering
\includegraphics[width=1.0\columnwidth]{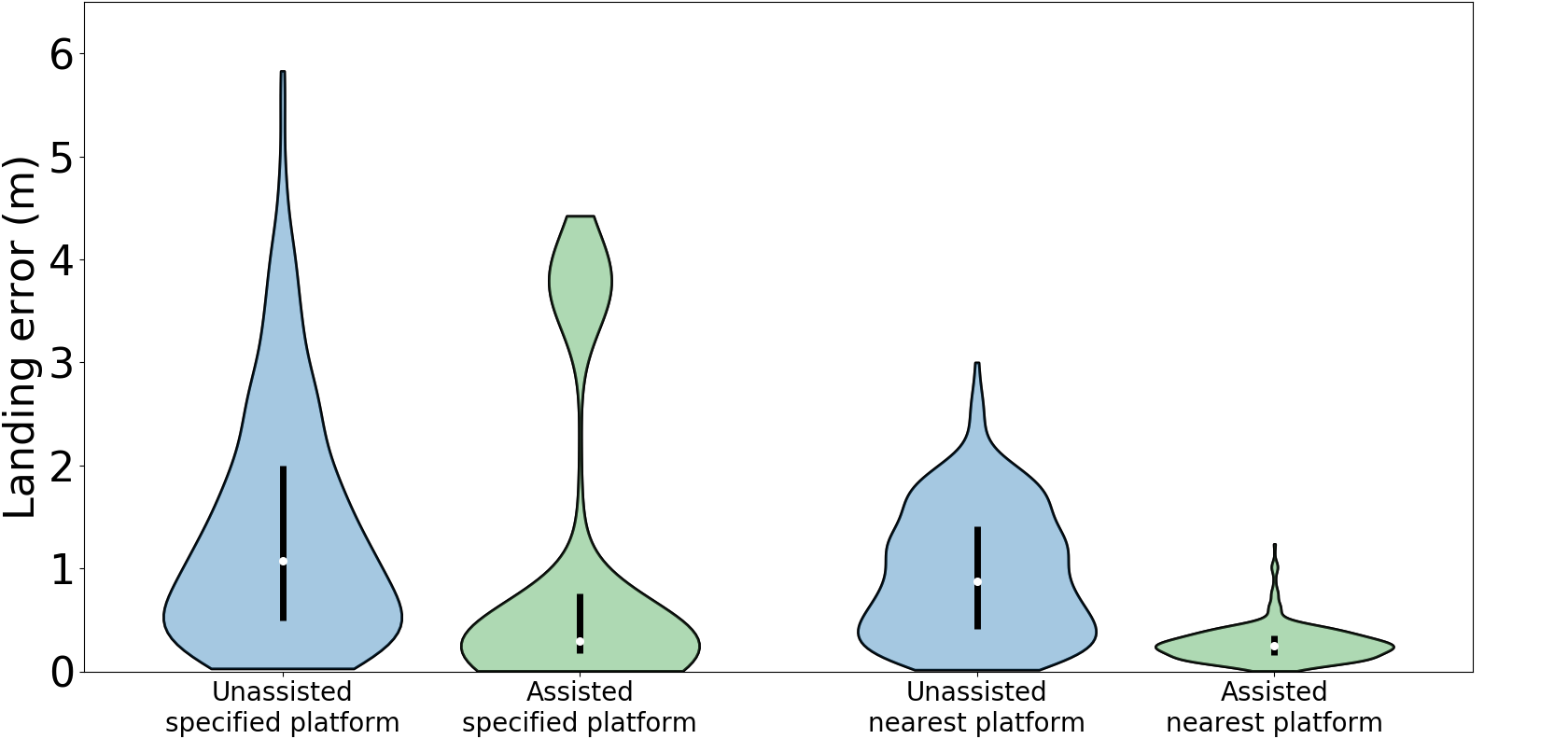}
\caption{Violin plot showing the distribution of errors whilst flying unassisted (blue) and assisted (green).}
\label{ViolinPlot}
\end{figure}

The distribution of pilot errors can be observed in Fig.~\ref{ViolinPlot}. Whilst flying assisted two distinct modes of errors are formed which coincide with the average distance between landing platforms of 4.0m. Due to the difficulty of the prescribed task, particularly with depth perception, novice pilots often attempted to land in the centre of the valley between platforms. This led to ambiguities in the pilot’s goal where the assistant correctly inferred the goal at a rate of 79.34\%. Of the correctly inferred goals, 93.34\% led to a successful landing. 

Participants gave an average SUS rating of 60.91 when unassisted,  which falls within the 32nd percentile based on \cite{SUSPercentiles}. Assisted participants gave an average SUS rating of 73.41, corresponding to the 68th percentile, an improvement of the perceived usability of the task by a factor of two. The results of the final survey can be viewed in Table.~\ref{UserStudyDemoTable} \&~\ref{SurveyTable2}. Participants preferred the assisted condition as they felt more confident in their ability to accurately land, and perceived a reduction in the difficulty of the task. Participants were also given the opportunity to provide additional feedback about the study; 24 out of 33 participants submitted optional comments. 

\def\VarOA{45}
\def\VarOB{55}
\def\VarPA{76}
\def\VarPB{6}
\def\VarPC{9}
\def\VarPD{3}
\def\VarPE{6}
\def\VarQA{52}
\def\VarQB{12}
\def\VarQC{24}
\def\VarQD{12}
\def\VarQE{0}
\def\VarRA{18}
\def\VarRB{27}
\def\VarRC{9}
\def\VarRD{18}
\def\VarRE{27}
\def\VarSA{48}
\def\VarSB{6}
\def\VarSC{18}
\def\VarSD{12}
\def\VarSE{15}
\def\VarTA{45}
\def\VarTB{3}
\def\VarTC{6}
\def\VarTD{3}
\def\VarTE{12}
\def\VarUA{76}
\def\VarUB{67}
\def\VarUC{61}
\def\VarUD{24}
\def\VarUE{9}

\tikzset{ 
    tableD/.style={
        matrix of nodes,
        nodes={
            rectangle,
            draw=white,
            align=center,
            text width=2.7em
        },
        minimum height=1.5em,
        text depth=0.5em,
        text height=0.5em,
        font=\footnotesize,
        nodes in empty cells,
        column 1/.style={
            nodes={text width=6.8em, align=justify}
        }
    }
}

 \newcommand{\DrawCell}[4] 
  {
  \def\TL{#1}
  \def\TR{#2}
  \def\BR{#3}
  \def\BL{#4}
  \draw (\TL) -- (\TR) -- (\BR) -- (\BL) -- (\TL);
  }
  
  \newcommand{\DrawBottomlessCell}[4]
  {
  \def\TL{#1}
  \def\TR{#2}
  \def\BR{#3}
  \def\BL{#4}
  \draw (\BL)  -- (\TL) -- (\TR) -- (\BR);
  }
  
  \newcommand{\DrawToplessCell}[4] 
  {
  \def\TL{#1}
  \def\TR{#2}
  \def\BR{#3}
  \def\BL{#4}
  \draw (\TL)  -- (\BL) -- (\BR) -- (\TR);
  }

\begin{figure}
\captionof{table}{Final survey - Demographics}
\label{UserStudyDemoTable}
\begin{tikzpicture}[trim left=-4.4cm] 
\matrix[tableD] (Demo)
{
&&&&&\\
&&&&&\\
&&&&&\\
&&&&&\\
&&&&&\\
&&&&&\\
&&&&&\\
&&&&&\\
&&&&&\\
&&&&&\\
&&&&&\\
&&&&&\\
&&&&&\\
&&&&&\\
&&&&&\\
};
\definecolor{OddColor}{rgb}{0.122, 0.466, 0.706}
\definecolor{EvenColor}{rgb}{0.550, 0.792, 0.576}
\def\OPACITYODD{0.4}
\def\OPACITYEVEN{0.4}
\DrawBottomlessCell{Demo-1-1.north west}{Demo-1-1.north east}{Demo-1-1.south east}{Demo-1-1.south west}
\DrawToplessCell{Demo-2-1.north west}{Demo-2-1.north east}{Demo-2-1.south east}{Demo-2-1.south west}

\DrawBottomlessCell{Demo-3-1.north west}{Demo-3-1.north east}{Demo-3-1.south east}{Demo-3-1.south west}
\DrawToplessCell{Demo-4-1.north west}{Demo-4-1.north east}{Demo-4-1.south east}{Demo-4-1.south west}

\DrawBottomlessCell{Demo-5-1.north west}{Demo-5-1.north east}{Demo-5-1.south east}{Demo-5-1.south west}
\DrawToplessCell{Demo-6-1.north west}{Demo-6-1.north east}{Demo-6-1.south east}{Demo-6-1.south west}

\DrawBottomlessCell{Demo-7-1.north west}{Demo-7-1.north east}{Demo-7-1.south east}{Demo-7-1.south west}
\DrawToplessCell{Demo-8-1.north west}{Demo-8-1.north east}{Demo-8-1.south east}{Demo-8-1.south west}

\DrawBottomlessCell{Demo-9-1.north west}{Demo-9-1.north east}{Demo-9-1.south east}{Demo-9-1.south west}
\DrawToplessCell{Demo-10-1.north west}{Demo-10-1.north east}{Demo-10-1.south east}{Demo-10-1.south west}

\DrawBottomlessCell{Demo-11-1.north west}{Demo-11-1.north east}{Demo-11-1.south east}{Demo-11-1.south west}
\DrawToplessCell{Demo-12-1.north west}{Demo-12-1.north east}{Demo-12-1.south east}{Demo-12-1.south west}

\DrawBottomlessCell{Demo-13-1.north west}{Demo-13-1.north east}{Demo-13-1.south east}{Demo-13-1.south west}
\DrawToplessCell{Demo-14-1.north west}{Demo-14-1.north east}{Demo-14-1.south east}{Demo-14-1.south west}
%
\node[anchor = north west, align=justify, text width=6.8em, font=\scriptsize] at (Demo-1-1.north west) {\textbf{1}: Have you had any experience with flying a physical drone?};

\node[anchor = north west, align=justify, text width=6.8em, font=\scriptsize] at (Demo-3-1.north west) {\textbf{2}: How many hours of experience have you with flying physical drones?};

\node[anchor = north west, align=justify, text width=6.8em, font=\scriptsize] at (Demo-5-1.north west) {\textbf{3}: How confident are you with flying physical drones?};

\node[anchor = north west, align=justify, text width=6.8em, font=\scriptsize] at (Demo-7-1.north west) {\textbf{4}: For what purpose have you flown a physical drone for? (Tick all that may apply)};

\node[anchor = north west, align=justify, text width=6.8em, font=\scriptsize] at (Demo-9-1.north west) {\textbf{5}: How often do you play video games? (Eg: PC/PS4/Xbox)};

\node[anchor = north west, align=justify, text width=6.8em, font=\scriptsize] at (Demo-11-1.north west) {\textbf{6}: How many hours have you spent on games involving flying?};

\node[anchor = north west, align=justify, text width=6.8em, font=\scriptsize] at (Demo-13-1.north west) {\textbf{7}: When you play video games, what input device do you use? (Tick all that may apply)};
%
\DrawCell{Demo-1-2.north west}{Demo-1-4.north}{Demo-1-4.south}{Demo-1-2.south west}
\DrawCell{Demo-1-4.north}{Demo-1-6.north east}{Demo-1-6.south east}{Demo-1-4.south}

\DrawCell{Demo-2-2.north west}{Demo-2-4.north}{Demo-2-4.south}{Demo-2-2.south west}
\DrawCell{Demo-2-4.north}{Demo-2-6.north east}{Demo-2-6.south east}{Demo-2-4.south}
\node[anchor = north, align=justify, text width=2.8em, font=\footnotesize] at (Demo-1-3.north) {\\ Yes};
\node[anchor = north west, align=justify, text width=2.8em, font=\footnotesize] at (Demo-1-5.north) {\\ No};
\node[anchor = north, align=justify, text width=2.8em, font=\footnotesize] at (Demo-2-3.north) {\\ \textbf{\VarOA \%}};
\node[anchor = north west, align=justify, text width=2.8em, font=\footnotesize] at (Demo-2-5.north) {\\ \textbf{\VarOB \%}};
\draw[opacity=\OPACITYODD,fill=OddColor] ($ (Demo-2-4.north)!1 - \VarOA / 100!(Demo-2-4.south) $) rectangle(Demo-2-2.south west);
\draw[opacity=\OPACITYODD,fill=OddColor] ($ (Demo-2-6.north east)!1 - \VarOB / 100!(Demo-2-6.south east) $) rectangle(Demo-2-4.south);

\DrawCell{Demo-3-2.north west}{Demo-3-2.north east}{Demo-3-2.south east}{Demo-3-2.south west}
\DrawCell{Demo-3-3.north west}{Demo-3-3.north east}{Demo-3-3.south east}{Demo-3-3.south west}
\DrawCell{Demo-3-4.north west}{Demo-3-4.north east}{Demo-3-4.south east}{Demo-3-4.south west}
\DrawCell{Demo-3-5.north west}{Demo-3-5.north east}{Demo-3-5.south east}{Demo-3-5.south west}
\DrawCell{Demo-3-6.north west}{Demo-3-6.north east}{Demo-3-6.south east}{Demo-3-6.south west}

\DrawCell{Demo-4-2.north west}{Demo-4-2.north east}{Demo-4-2.south east}{Demo-4-2.south west}
\DrawCell{Demo-4-3.north west}{Demo-4-3.north east}{Demo-4-3.south east}{Demo-4-3.south west}
\DrawCell{Demo-4-4.north west}{Demo-4-4.north east}{Demo-4-4.south east}{Demo-4-4.south west}
\DrawCell{Demo-4-5.north west}{Demo-4-5.north east}{Demo-4-5.south east}{Demo-4-5.south west}
\DrawCell{Demo-4-6.north west}{Demo-4-6.north east}{Demo-4-6.south east}{Demo-4-6.south west}
\node[anchor = north west, align=center, text width=2.8em, font=\tiny] at (Demo-3-2.north west) {\\ 0-5};
\node[anchor = north west, align=center, text width=2.8em, font=\tiny] at (Demo-3-3.north west) {\\ 5-10};
\node[anchor = north west, align=center, text width=2.8em, font=\tiny] at (Demo-3-4.north west) {\\ 10-20};
\node[anchor = north west, align=center, text width=2.8em, font=\tiny] at (Demo-3-5.north west) {\\ 20-50};
\node[anchor = north west, align=center, text width=2.8em, font=\tiny] at (Demo-3-6.north west) {\\ 50+};
\node[anchor = north west, align=center, text width=2.8em, font=\scriptsize] at (Demo-4-2.north west) {\\ \textbf{\VarPA \%}};
\node[anchor = north west, align=center, text width=2.8em, font=\scriptsize] at (Demo-4-3.north west) {\\ \textbf{\VarPB \%}};
\node[anchor = north west, align=center, text width=2.8em, font=\scriptsize] at (Demo-4-4.north west) {\\ \textbf{\VarPC \%}};
\node[anchor = north west, align=center, text width=2.8em, font=\scriptsize] at (Demo-4-5.north west) {\\ \textbf{\VarPD \%}};
\node[anchor = north west, align=center, text width=2.8em, font=\scriptsize] at (Demo-4-6.north west) {\\ \textbf{\VarPE \%}};
\draw[opacity=\OPACITYEVEN,fill=EvenColor] ($ (Demo-4-2.north west)!1 - \VarPA / 100!(Demo-4-2.south west) $) rectangle(Demo-4-2.south east);
\draw[opacity=\OPACITYEVEN,fill=EvenColor] ($ (Demo-4-3.north west)!1 - \VarPB / 100!(Demo-4-3.south west) $) rectangle(Demo-4-3.south east);
\draw[opacity=\OPACITYEVEN,fill=EvenColor] ($ (Demo-4-4.north west)!1 - \VarPC / 100!(Demo-4-4.south west) $) rectangle(Demo-4-4.south east);
\draw[opacity=\OPACITYEVEN,fill=EvenColor] ($ (Demo-4-5.north west)!1 - \VarPD / 100!(Demo-4-5.south west) $) rectangle(Demo-4-5.south east);
\draw[opacity=\OPACITYEVEN,fill=EvenColor] ($ (Demo-4-6.north west)!1 - \VarPE / 100!(Demo-4-6.south west) $) rectangle(Demo-4-6.south east);

\DrawCell{Demo-5-2.north west}{Demo-5-2.north east}{Demo-5-2.south east}{Demo-5-2.south west}
\DrawCell{Demo-5-3.north west}{Demo-5-3.north east}{Demo-5-3.south east}{Demo-5-3.south west}
\DrawCell{Demo-5-4.north west}{Demo-5-4.north east}{Demo-5-4.south east}{Demo-5-4.south west}
\DrawCell{Demo-5-5.north west}{Demo-5-5.north east}{Demo-5-5.south east}{Demo-5-5.south west}
\DrawCell{Demo-5-6.north west}{Demo-5-6.north east}{Demo-5-6.south east}{Demo-5-6.south west}

\DrawCell{Demo-6-2.north west}{Demo-6-2.north east}{Demo-6-2.south east}{Demo-6-2.south west}
\DrawCell{Demo-6-3.north west}{Demo-6-3.north east}{Demo-6-3.south east}{Demo-6-3.south west}
\DrawCell{Demo-6-4.north west}{Demo-6-4.north east}{Demo-6-4.south east}{Demo-6-4.south west}
\DrawCell{Demo-6-5.north west}{Demo-6-5.north east}{Demo-6-5.south east}{Demo-6-5.south west}
\DrawCell{Demo-6-6.north west}{Demo-6-6.north east}{Demo-6-6.south east}{Demo-6-6.south west}
\node[anchor = north west, align=center, text width=2.8em, font=\tiny] at (Demo-5-2.north west) {Not confident \\ 1};
\node[anchor = north west, align=center, text width=2.8em, font=\tiny] at (Demo-5-3.north west) {\\ 2};
\node[anchor = north west, align=center, text width=2.8em, font=\tiny] at (Demo-5-4.north west) {\\ 3};
\node[anchor = north west, align=center, text width=2.8em, font=\tiny] at (Demo-5-5.north west) {\\ 4};
\node[anchor = north west, align=center, text width=2.8em, font=\tiny] at (Demo-5-6.north west) {Confident \\ 5};
\node[anchor = north west, align=center, text width=2.8em, font=\scriptsize] at (Demo-6-2.north west) {\\ \textbf{\VarQA \%}};
\node[anchor = north west, align=center, text width=2.8em, font=\scriptsize] at (Demo-6-3.north west) {\\ \textbf{\VarQB \%}};
\node[anchor = north west, align=center, text width=2.8em, font=\scriptsize] at (Demo-6-4.north west) {\\ \textbf{\VarQC \%}};
\node[anchor = north west, align=center, text width=2.8em, font=\scriptsize] at (Demo-6-5.north west) {\\ \textbf{\VarQD \%}};
\node[anchor = north west, align=center, text width=2.8em, font=\scriptsize] at (Demo-6-6.north west) {\\ \textbf{\VarQE \%}};
\draw[opacity=\OPACITYODD,fill=OddColor] ($ (Demo-6-2.north west)!1 - \VarQA / 100!(Demo-6-2.south west) $) rectangle(Demo-6-2.south east);
\draw[opacity=\OPACITYODD,fill=OddColor] ($ (Demo-6-3.north west)!1 - \VarQB / 100!(Demo-6-3.south west) $) rectangle(Demo-6-3.south east);
\draw[opacity=\OPACITYODD,fill=OddColor] ($ (Demo-6-4.north west)!1 - \VarQC / 100!(Demo-6-4.south west) $) rectangle(Demo-6-4.south east);
\draw[opacity=\OPACITYODD,fill=OddColor] ($ (Demo-6-5.north west)!1 - \VarQD / 100!(Demo-6-5.south west) $) rectangle(Demo-6-5.south east);
\draw[opacity=\OPACITYODD,fill=OddColor] ($ (Demo-6-6.north west)!1 - \VarQE / 100!(Demo-6-6.south west) $) rectangle(Demo-6-6.south east);

\DrawCell{Demo-7-2.north west}{Demo-7-2.north east}{Demo-7-2.south east}{Demo-7-2.south west}
\DrawCell{Demo-7-3.north west}{Demo-7-3.north east}{Demo-7-3.south east}{Demo-7-3.south west}
\DrawCell{Demo-7-4.north west}{Demo-7-4.north east}{Demo-7-4.south east}{Demo-7-4.south west}
\DrawCell{Demo-7-5.north west}{Demo-7-5.north east}{Demo-7-5.south east}{Demo-7-5.south west}
\DrawCell{Demo-7-6.north west}{Demo-7-6.north east}{Demo-7-6.south east}{Demo-7-6.south west}

\DrawCell{Demo-8-2.north west}{Demo-8-2.north east}{Demo-8-2.south east}{Demo-8-2.south west}
\DrawCell{Demo-8-3.north west}{Demo-8-3.north east}{Demo-8-3.south east}{Demo-8-3.south west}
\DrawCell{Demo-8-4.north west}{Demo-8-4.north east}{Demo-8-4.south east}{Demo-8-4.south west}
\DrawCell{Demo-8-5.north west}{Demo-8-5.north east}{Demo-8-5.south east}{Demo-8-5.south west}
\DrawCell{Demo-8-6.north west}{Demo-8-6.north east}{Demo-8-6.south east}{Demo-8-6.south west}
\node[anchor = north west, align=center, text width=2.8em, font=\tiny] at (Demo-7-2.north west) {\\ Recreational};
\node[anchor = north west, align=center, text width=2.8em, font=\tiny] at (Demo-7-3.north west) {Photography Filming};
\node[anchor = north west, align=center, text width=2.8em, font=\tiny] at (Demo-7-4.north west) {FPV Racing Aerobatics};
\node[anchor = north west, align=center, text width=2.8em, font=\tiny] at (Demo-7-5.north west) {\\ Inspection};
\node[anchor = north west, align=center, text width=2.8em, font=\tiny] at (Demo-7-6.north west) {\\ Other};
\node[anchor = north west, align=center, text width=2.8em, font=\scriptsize] at (Demo-8-2.north west) {\\ \textbf{\VarTA \%}};
\node[anchor = north west, align=center, text width=2.8em, font=\scriptsize] at (Demo-8-3.north west) {\\ \textbf{\VarTB \%}};
\node[anchor = north west, align=center, text width=2.8em, font=\scriptsize] at (Demo-8-4.north west) {\\ \textbf{\VarTC \%}};
\node[anchor = north west, align=center, text width=2.8em, font=\scriptsize] at (Demo-8-5.north west) {\\ \textbf{\VarTD \%}};
\node[anchor = north west, align=center, text width=2.8em, font=\scriptsize] at (Demo-8-6.north west) {\\ \textbf{\VarTE \%}};
\draw[opacity=\OPACITYEVEN,fill=EvenColor] ($ (Demo-8-2.north west)!1 - \VarTA / 100!(Demo-8-2.south west) $) rectangle(Demo-8-2.south east);
\draw[opacity=\OPACITYEVEN,fill=EvenColor] ($ (Demo-8-3.north west)!1 - \VarTB / 100!(Demo-8-3.south west) $) rectangle(Demo-8-3.south east);
\draw[opacity=\OPACITYEVEN,fill=EvenColor] ($ (Demo-8-4.north west)!1 - \VarTC / 100!(Demo-8-4.south west) $) rectangle(Demo-8-4.south east);
\draw[opacity=\OPACITYEVEN,fill=EvenColor] ($ (Demo-8-5.north west)!1 - \VarTD / 100!(Demo-8-5.south west) $) rectangle(Demo-8-5.south east);
\draw[opacity=\OPACITYEVEN,fill=EvenColor] ($ (Demo-8-6.north west)!1 - \VarTE / 100!(Demo-8-6.south west) $) rectangle(Demo-8-6.south east);

\DrawCell{Demo-9-2.north west}{Demo-9-2.north east}{Demo-9-2.south east}{Demo-9-2.south west}
\DrawCell{Demo-9-3.north west}{Demo-9-3.north east}{Demo-9-3.south east}{Demo-9-3.south west}
\DrawCell{Demo-9-4.north west}{Demo-9-4.north east}{Demo-9-4.south east}{Demo-9-4.south west}
\DrawCell{Demo-9-5.north west}{Demo-9-5.north east}{Demo-9-5.south east}{Demo-9-5.south west}
\DrawCell{Demo-9-6.north west}{Demo-9-6.north east}{Demo-9-6.south east}{Demo-9-6.south west}

\DrawCell{Demo-10-2.north west}{Demo-10-2.north east}{Demo-10-2.south east}{Demo-10-2.south west}
\DrawCell{Demo-10-3.north west}{Demo-10-3.north east}{Demo-10-3.south east}{Demo-10-3.south west}
\DrawCell{Demo-10-4.north west}{Demo-10-4.north east}{Demo-10-4.south east}{Demo-10-4.south west}
\DrawCell{Demo-10-5.north west}{Demo-10-5.north east}{Demo-10-5.south east}{Demo-10-5.south west}
\DrawCell{Demo-10-6.north west}{Demo-10-6.north east}{Demo-10-6.south east}{Demo-10-6.south west}
\node[anchor = north west, align=center, text width=2.8em, font=\tiny] at (Demo-9-2.north west) {\\ Never};
\node[anchor = north west, align=center, text width=2.8em, font=\tiny] at (Demo-9-3.north west) {\\ Monthly};
\node[anchor = north west, align=center, text width=2.8em, font=\tiny] at (Demo-9-4.north west) {\\ Weekly};
\node[anchor = north west, align=center, text width=2.8em, font=\tiny] at (Demo-9-5.north west) {\\ Regularly};
\node[anchor = north west, align=center, text width=2.8em, font=\tiny] at (Demo-9-6.north west) {\\ Daily};
\node[anchor = north west, align=center, text width=2.8em, font=\scriptsize] at (Demo-10-2.north west) {\\ \textbf{\VarRA \%}};
\node[anchor = north west, align=center, text width=2.8em, font=\scriptsize] at (Demo-10-3.north west) {\\ \textbf{\VarRB \%}};
\node[anchor = north west, align=center, text width=2.8em, font=\scriptsize] at (Demo-10-4.north west) {\\ \textbf{\VarRC \%}};
\node[anchor = north west, align=center, text width=2.8em, font=\scriptsize] at (Demo-10-5.north west) {\\ \textbf{\VarRD \%}};
\node[anchor = north west, align=center, text width=2.8em, font=\scriptsize] at (Demo-10-6.north west) {\\ \textbf{\VarRE \%}};
\draw[opacity=\OPACITYODD,fill=OddColor] ($ (Demo-10-2.north west)!1 - \VarRA / 100!(Demo-10-2.south west) $) rectangle(Demo-10-2.south east);
\draw[opacity=\OPACITYODD,fill=OddColor] ($ (Demo-10-3.north west)!1 - \VarRB / 100!(Demo-10-3.south west) $) rectangle(Demo-10-3.south east);
\draw[opacity=\OPACITYODD,fill=OddColor] ($ (Demo-10-4.north west)!1 - \VarRC / 100!(Demo-10-4.south west) $) rectangle(Demo-10-4.south east);
\draw[opacity=\OPACITYODD,fill=OddColor] ($ (Demo-10-5.north west)!1 - \VarRD / 100!(Demo-10-5.south west) $) rectangle(Demo-10-5.south east);
\draw[opacity=\OPACITYODD,fill=OddColor] ($ (Demo-10-6.north west)!1 - \VarRE / 100!(Demo-10-6.south west) $) rectangle(Demo-10-6.south east);

\DrawCell{Demo-11-2.north west}{Demo-11-2.north east}{Demo-11-2.south east}{Demo-11-2.south west}
\DrawCell{Demo-11-3.north west}{Demo-11-3.north east}{Demo-11-3.south east}{Demo-11-3.south west}
\DrawCell{Demo-11-4.north west}{Demo-11-4.north east}{Demo-11-4.south east}{Demo-11-4.south west}
\DrawCell{Demo-11-5.north west}{Demo-11-5.north east}{Demo-11-5.south east}{Demo-11-5.south west}
\DrawCell{Demo-11-6.north west}{Demo-11-6.north east}{Demo-11-6.south east}{Demo-11-6.south west}

\DrawCell{Demo-12-2.north west}{Demo-12-2.north east}{Demo-12-2.south east}{Demo-12-2.south west}
\DrawCell{Demo-12-3.north west}{Demo-12-3.north east}{Demo-12-3.south east}{Demo-12-3.south west}
\DrawCell{Demo-12-4.north west}{Demo-12-4.north east}{Demo-12-4.south east}{Demo-12-4.south west}
\DrawCell{Demo-12-5.north west}{Demo-12-5.north east}{Demo-12-5.south east}{Demo-12-5.south west}
\DrawCell{Demo-12-6.north west}{Demo-12-6.north east}{Demo-12-6.south east}{Demo-12-6.south west}
\node[anchor = north west, align=center, text width=2.8em, font=\tiny] at (Demo-11-2.north west) {\\ 0-10};
\node[anchor = north west, align=center, text width=2.8em, font=\tiny] at (Demo-11-3.north west) {\\ 10-25};
\node[anchor = north west, align=center, text width=2.8em, font=\tiny] at (Demo-11-4.north west) {\\ 25-100};
\node[anchor = north west, align=center, text width=2.8em, font=\tiny] at (Demo-11-5.north west) {\\ 100-200};
\node[anchor = north west, align=center, text width=2.8em, font=\tiny] at (Demo-11-6.north west) {\\ 200+};
\node[anchor = north west, align=center, text width=2.8em, font=\scriptsize] at (Demo-12-2.north west) {\\ \textbf{\VarSA \%}};
\node[anchor = north west, align=center, text width=2.8em, font=\scriptsize] at (Demo-12-3.north west) {\\ \textbf{\VarSB \%}};
\node[anchor = north west, align=center, text width=2.8em, font=\scriptsize] at (Demo-12-4.north west) {\\ \textbf{\VarSC \%}};
\node[anchor = north west, align=center, text width=2.8em, font=\scriptsize] at (Demo-12-5.north west) {\\ \textbf{\VarSD \%}};
\node[anchor = north west, align=center, text width=2.8em, font=\scriptsize] at (Demo-12-6.north west) {\\ \textbf{\VarSE \%}};
\draw[opacity=\OPACITYEVEN,fill=EvenColor] ($ (Demo-12-2.north west)!1 - \VarSA / 100!(Demo-12-2.south west) $) rectangle(Demo-12-2.south east);
\draw[opacity=\OPACITYEVEN,fill=EvenColor] ($ (Demo-12-3.north west)!1 - \VarSB / 100!(Demo-12-3.south west) $) rectangle(Demo-12-3.south east);
\draw[opacity=\OPACITYEVEN,fill=EvenColor] ($ (Demo-12-4.north west)!1 - \VarSC / 100!(Demo-12-4.south west) $) rectangle(Demo-12-4.south east);
\draw[opacity=\OPACITYEVEN,fill=EvenColor] ($ (Demo-12-5.north west)!1 - \VarSD / 100!(Demo-12-5.south west) $) rectangle(Demo-12-5.south east);
\draw[opacity=\OPACITYEVEN,fill=EvenColor] ($ (Demo-12-6.north west)!1 - \VarSE / 100!(Demo-12-6.south west) $) rectangle(Demo-12-6.south east);

\DrawCell{Demo-13-2.north west}{Demo-13-2.north east}{Demo-13-2.south east}{Demo-13-2.south west}
\DrawCell{Demo-13-3.north west}{Demo-13-3.north east}{Demo-13-3.south east}{Demo-13-3.south west}
\DrawCell{Demo-13-4.north west}{Demo-13-4.north east}{Demo-13-4.south east}{Demo-13-4.south west}
\DrawCell{Demo-13-5.north west}{Demo-13-5.north east}{Demo-13-5.south east}{Demo-13-5.south west}
\DrawCell{Demo-13-6.north west}{Demo-13-6.north east}{Demo-13-6.south east}{Demo-13-6.south west}

\DrawCell{Demo-14-2.north west}{Demo-14-2.north east}{Demo-14-2.south east}{Demo-14-2.south west}
\DrawCell{Demo-14-3.north west}{Demo-14-3.north east}{Demo-14-3.south east}{Demo-14-3.south west}
\DrawCell{Demo-14-4.north west}{Demo-14-4.north east}{Demo-14-4.south east}{Demo-14-4.south west}
\DrawCell{Demo-14-5.north west}{Demo-14-5.north east}{Demo-14-5.south east}{Demo-14-5.south west}
\DrawCell{Demo-14-6.north west}{Demo-14-6.north east}{Demo-14-6.south east}{Demo-14-6.south west}
\node[anchor = north west, align=center, text width=2.8em, font=\tiny] at (Demo-13-2.north west) {\\ Keyboard};
\node[anchor = north west, align=center, text width=2.8em, font=\tiny] at (Demo-13-3.north west) {\\ Mouse};
\node[anchor = north west, align=center, text width=2.8em, font=\tiny] at (Demo-13-4.north west) {Hand held controller};
\node[anchor = north west, align=center, text width=2.8em, font=\tiny] at (Demo-13-5.north west) {\\ Touch screen};
\node[anchor = north west, align=center, text width=2.8em, font=\tiny] at (Demo-13-6.north west) {\\ Other};
\node[anchor = north west, align=center, text width=2.8em, font=\scriptsize] at (Demo-14-2.north west) {\\ \textbf{\VarUA \%}};
\node[anchor = north west, align=center, text width=2.8em, font=\scriptsize] at (Demo-14-3.north west) {\\ \textbf{\VarUB \%}};
\node[anchor = north west, align=center, text width=2.8em, font=\scriptsize] at (Demo-14-4.north west) {\\ \textbf{\VarUC \%}};
\node[anchor = north west, align=center, text width=2.8em, font=\scriptsize] at (Demo-14-5.north west) {\\ \textbf{\VarUD \%}};
\node[anchor = north west, align=center, text width=2.8em, font=\scriptsize] at (Demo-14-6.north west) {\\ \textbf{\VarUE \%}};
\draw[opacity=\OPACITYODD,fill=OddColor] ($ (Demo-14-2.north west)!1 - \VarUA / 100!(Demo-14-2.south west) $) rectangle(Demo-14-2.south east);
\draw[opacity=\OPACITYODD,fill=OddColor] ($ (Demo-14-3.north west)!1 - \VarUB / 100!(Demo-14-3.south west) $) rectangle(Demo-14-3.south east);
\draw[opacity=\OPACITYODD,fill=OddColor] ($ (Demo-14-4.north west)!1 - \VarUC / 100!(Demo-14-4.south west) $) rectangle(Demo-14-4.south east);
\draw[opacity=\OPACITYODD,fill=OddColor] ($ (Demo-14-5.north west)!1 - \VarUD / 100!(Demo-14-5.south west) $) rectangle(Demo-14-5.south east);
\draw[opacity=\OPACITYODD,fill=OddColor] ($ (Demo-14-6.north west)!1 - \VarUE / 100!(Demo-14-6.south west) $) rectangle(Demo-14-6.south east);
\end{tikzpicture} 
\end{figure}

\def\VarAA{21}
\def\VarAB{42}
\def\VarAC{15}
\def\VarAD{18}
\def\VarAE{3}
\def\VarBA{6}
\def\VarBB{9}
\def\VarBC{18}
\def\VarBD{58}
\def\VarBE{9}
\def\VarCA{9}
\def\VarCB{18}
\def\VarCC{27}
\def\VarCD{39}
\def\VarCE{6}
\def\VarDA{24}
\def\VarDB{36}
\def\VarDC{21}
\def\VarDD{15}
\def\VarDE{3}
\def\VarEA{6}
\def\VarEB{18}
\def\VarEC{6}
\def\VarED{39}
\def\VarEE{30}
\def\VarFA{15}
\def\VarFB{36}
\def\VarFC{24}
\def\VarFD{21}
\def\VarFE{3}
\def\VarGA{3}
\def\VarGB{9}
\def\VarGC{18}
\def\VarGD{30}
\def\VarGE{39}
\def\VarHA{9}
\def\VarHB{15}
\def\VarHC{21}
\def\VarHD{18}
\def\VarHE{36}
\def\VarIA{9}
\def\VarIB{18}
\def\VarIC{33}
\def\VarID{33}
\def\VarIE{6}
\def\VarJA{9}
\def\VarJB{33}
\def\VarJC{24}
\def\VarJD{30}
\def\VarJE{3}
\def\VarKA{18}
\def\VarKB{36}
\def\VarKC{15}
\def\VarKD{12}
\def\VarKE{18}
\def\VarLA{0}
\def\VarLB{6}
\def\VarLC{24}
\def\VarLD{39}
\def\VarLE{30}
\def\VarMA{0}
\def\VarMB{24}
\def\VarMC{30}
\def\VarMD{24}
\def\VarME{21}
\def\VarNA{12}
\def\VarNB{27}
\def\VarNC{18}
\def\VarND{24}
\def\VarNE{18}

\tikzset{ 
    table/.style={
        matrix of nodes,
        nodes={
            rectangle,
            draw=white,
            align=center,
        },
        minimum height=1.5em,
        text depth=1.0em,
        text height=1.0ex,
        font=\footnotesize,
        nodes in empty cells,
        column 1/.style={
            nodes={text width=12.6em, align=justify}
        },
        row 1/.style={
            nodes={
                font=\bfseries, align=center, text depth=0.05em, text height=1.2ex, minimum height=0.2em
            }
        },
        row 15/.style={
            nodes={
                text depth=2.0em
            }
        }
    }
}

\begin{figure}
\captionof{table}{Final survey - User experience }
\label{SurveyTable2}
\begin{tikzpicture}
\matrix [table,text width=1.5em] (M)
{
 &1&2&3&4&5\\
\textbf{8}: I felt confident in landing unassisted&&&&&\\
\textbf{9}: I felt confident in landing assisted&&&&&\\
\textbf{10}:I felt stressed when landing unassisted&&&&&\\
\textbf{11}: I felt stressed when landing assisted&&&&&\\
\textbf{12}: I felt the task was difficult to perform unassisted&&&&&\\
\textbf{13}: I felt the task was difficult to perform assisted&&&&&\\
\textbf{14}: I was able to land with greater accuracy whilst being assisted&&&&&\\
\textbf{15}: I was able to land quicker whilst being assisted&&&&&\\
\textbf{16}: I trust the actions of the assistant&&&&&\\
\textbf{17}: The assistant didn't do what I wanted to do&&&&&\\
\textbf{18}: I had to fight the assistant for control&&&&&\\
\textbf{19}: I prefer to use the assistant when landing&&&&&\\
\textbf{20}: I felt being unassisted gave me more freedom&&&&&\\
\textbf{21}: I believe that if I practiced, I could perform better without assistance than with assistance&&&&&\\
};

\definecolor{OddColor}{rgb}{0.122, 0.466, 0.706}
\definecolor{EvenColor}{rgb}{0.550, 0.792, 0.576}
\def\OPACITYODD{0.4}
\def\OPACITYEVEN{0.4}

\DrawCell{M-1-1.north west}{M-1-1.north east}{M-1-1.south east}{M-1-1.south west}
\DrawCell{M-1-2.north west}{M-1-2.north east}{M-1-2.south east}{M-1-2.south west}
\DrawCell{M-1-3.north west}{M-1-3.north east}{M-1-3.south east}{M-1-3.south west}
\DrawCell{M-1-4.north west}{M-1-4.north east}{M-1-4.south east}{M-1-4.south west}
\DrawCell{M-1-5.north west}{M-1-5.north east}{M-1-5.south east}{M-1-5.south west}
\DrawCell{M-1-6.north west}{M-1-6.north east}{M-1-6.south east}{M-1-6.south west}

\DrawCell{M-2-1.north west}{M-2-1.north east}{M-2-1.south east}{M-2-1.south west}
\DrawCell{M-2-2.north west}{M-2-2.north east}{M-2-2.south east}{M-2-2.south west}
\DrawCell{M-2-3.north west}{M-2-3.north east}{M-2-3.south east}{M-2-3.south west}
\DrawCell{M-2-4.north west}{M-2-4.north east}{M-2-4.south east}{M-2-4.south west}
\DrawCell{M-2-5.north west}{M-2-5.north east}{M-2-5.south east}{M-2-5.south west}
\DrawCell{M-2-6.north west}{M-2-6.north east}{M-2-6.south east}{M-2-6.south west}

\DrawCell{M-3-1.north west}{M-3-1.north east}{M-3-1.south east}{M-3-1.south west}
\DrawCell{M-3-2.north west}{M-3-2.north east}{M-3-2.south east}{M-3-2.south west}
\DrawCell{M-3-3.north west}{M-3-3.north east}{M-3-3.south east}{M-3-3.south west}
\DrawCell{M-3-4.north west}{M-3-4.north east}{M-3-4.south east}{M-3-4.south west}
\DrawCell{M-3-5.north west}{M-3-5.north east}{M-3-5.south east}{M-3-5.south west}
\DrawCell{M-3-6.north west}{M-3-6.north east}{M-3-6.south east}{M-3-6.south west}

\DrawCell{M-4-1.north west}{M-4-1.north east}{M-4-1.south east}{M-4-1.south west}
\DrawCell{M-4-2.north west}{M-4-2.north east}{M-4-2.south east}{M-4-2.south west}
\DrawCell{M-4-3.north west}{M-4-3.north east}{M-4-3.south east}{M-4-3.south west}
\DrawCell{M-4-4.north west}{M-4-4.north east}{M-4-4.south east}{M-4-4.south west}
\DrawCell{M-4-5.north west}{M-4-5.north east}{M-4-5.south east}{M-4-5.south west}
\DrawCell{M-4-6.north west}{M-4-6.north east}{M-4-6.south east}{M-4-6.south west}

\DrawCell{M-5-1.north west}{M-5-1.north east}{M-5-1.south east}{M-5-1.south west}
\DrawCell{M-5-2.north west}{M-5-2.north east}{M-5-2.south east}{M-5-2.south west}
\DrawCell{M-5-3.north west}{M-5-3.north east}{M-5-3.south east}{M-5-3.south west}
\DrawCell{M-5-4.north west}{M-5-4.north east}{M-5-4.south east}{M-5-4.south west}
\DrawCell{M-5-5.north west}{M-5-5.north east}{M-5-5.south east}{M-5-5.south west}
\DrawCell{M-5-6.north west}{M-5-6.north east}{M-5-6.south east}{M-5-6.south west}

\DrawCell{M-6-1.north west}{M-6-1.north east}{M-6-1.south east}{M-6-1.south west}
\DrawCell{M-6-2.north west}{M-6-2.north east}{M-6-2.south east}{M-6-2.south west}
\DrawCell{M-6-3.north west}{M-6-3.north east}{M-6-3.south east}{M-6-3.south west}
\DrawCell{M-6-4.north west}{M-6-4.north east}{M-6-4.south east}{M-6-4.south west}
\DrawCell{M-6-5.north west}{M-6-5.north east}{M-6-5.south east}{M-6-5.south west}
\DrawCell{M-6-6.north west}{M-6-6.north east}{M-6-6.south east}{M-6-6.south west}

\DrawCell{M-7-1.north west}{M-7-1.north east}{M-7-1.south east}{M-7-1.south west}
\DrawCell{M-7-2.north west}{M-7-2.north east}{M-7-2.south east}{M-7-2.south west}
\DrawCell{M-7-3.north west}{M-7-3.north east}{M-7-3.south east}{M-7-3.south west}
\DrawCell{M-7-4.north west}{M-7-4.north east}{M-7-4.south east}{M-7-4.south west}
\DrawCell{M-7-5.north west}{M-7-5.north east}{M-7-5.south east}{M-7-5.south west}
\DrawCell{M-7-6.north west}{M-7-6.north east}{M-7-6.south east}{M-7-6.south west}

\DrawCell{M-8-1.north west}{M-8-1.north east}{M-8-1.south east}{M-8-1.south west}
\DrawCell{M-8-2.north west}{M-8-2.north east}{M-8-2.south east}{M-8-2.south west}
\DrawCell{M-8-3.north west}{M-8-3.north east}{M-8-3.south east}{M-8-3.south west}
\DrawCell{M-8-4.north west}{M-8-4.north east}{M-8-4.south east}{M-8-4.south west}
\DrawCell{M-8-5.north west}{M-8-5.north east}{M-8-5.south east}{M-8-5.south west}
\DrawCell{M-8-6.north west}{M-8-6.north east}{M-8-6.south east}{M-8-6.south west}

\DrawCell{M-9-1.north west}{M-9-1.north east}{M-9-1.south east}{M-9-1.south west}
\DrawCell{M-9-2.north west}{M-9-2.north east}{M-9-2.south east}{M-9-2.south west}
\DrawCell{M-9-3.north west}{M-9-3.north east}{M-9-3.south east}{M-9-3.south west}
\DrawCell{M-9-4.north west}{M-9-4.north east}{M-9-4.south east}{M-9-4.south west}
\DrawCell{M-9-5.north west}{M-9-5.north east}{M-9-5.south east}{M-9-5.south west}
\DrawCell{M-9-6.north west}{M-9-6.north east}{M-9-6.south east}{M-9-6.south west}

\DrawCell{M-10-1.north west}{M-10-1.north east}{M-10-1.south east}{M-10-1.south west}
\DrawCell{M-10-2.north west}{M-10-2.north east}{M-10-2.south east}{M-10-2.south west}
\DrawCell{M-10-3.north west}{M-10-3.north east}{M-10-3.south east}{M-10-3.south west}
\DrawCell{M-10-4.north west}{M-10-4.north east}{M-10-4.south east}{M-10-4.south west}
\DrawCell{M-10-5.north west}{M-10-5.north east}{M-10-5.south east}{M-10-5.south west}
\DrawCell{M-10-6.north west}{M-10-6.north east}{M-10-6.south east}{M-10-6.south west}

\DrawCell{M-11-1.north west}{M-11-1.north east}{M-11-1.south east}{M-11-1.south west}
\DrawCell{M-11-2.north west}{M-11-2.north east}{M-11-2.south east}{M-11-2.south west}
\DrawCell{M-11-3.north west}{M-11-3.north east}{M-11-3.south east}{M-11-3.south west}
\DrawCell{M-11-4.north west}{M-11-4.north east}{M-11-4.south east}{M-11-4.south west}
\DrawCell{M-11-5.north west}{M-11-5.north east}{M-11-5.south east}{M-11-5.south west}
\DrawCell{M-11-6.north west}{M-11-6.north east}{M-11-6.south east}{M-11-6.south west}

\DrawCell{M-12-1.north west}{M-12-1.north east}{M-12-1.south east}{M-12-1.south west}
\DrawCell{M-12-2.north west}{M-12-2.north east}{M-12-2.south east}{M-12-2.south west}
\DrawCell{M-12-3.north west}{M-12-3.north east}{M-12-3.south east}{M-12-3.south west}
\DrawCell{M-12-4.north west}{M-12-4.north east}{M-12-4.south east}{M-12-4.south west}
\DrawCell{M-12-5.north west}{M-12-5.north east}{M-12-5.south east}{M-12-5.south west}
\DrawCell{M-12-6.north west}{M-12-6.north east}{M-12-6.south east}{M-12-6.south west}

\DrawCell{M-13-1.north west}{M-13-1.north east}{M-13-1.south east}{M-13-1.south west}
\DrawCell{M-13-2.north west}{M-13-2.north east}{M-13-2.south east}{M-13-2.south west}
\DrawCell{M-13-3.north west}{M-13-3.north east}{M-13-3.south east}{M-13-3.south west}
\DrawCell{M-13-4.north west}{M-13-4.north east}{M-13-4.south east}{M-13-4.south west}
\DrawCell{M-13-5.north west}{M-13-5.north east}{M-13-5.south east}{M-13-5.south west}
\DrawCell{M-13-6.north west}{M-13-6.north east}{M-13-6.south east}{M-13-6.south west}

\DrawCell{M-14-1.north west}{M-14-1.north east}{M-14-1.south east}{M-14-1.south west}
\DrawCell{M-14-2.north west}{M-14-2.north east}{M-14-2.south east}{M-14-2.south west}
\DrawCell{M-14-3.north west}{M-14-3.north east}{M-14-3.south east}{M-14-3.south west}
\DrawCell{M-14-4.north west}{M-14-4.north east}{M-14-4.south east}{M-14-4.south west}
\DrawCell{M-14-5.north west}{M-14-5.north east}{M-14-5.south east}{M-14-5.south west}
\DrawCell{M-14-6.north west}{M-14-6.north east}{M-14-6.south east}{M-14-6.south west}

\DrawCell{M-15-1.north west}{M-15-1.north east}{M-15-1.south east}{M-15-1.south west}
\DrawCell{M-15-2.north west}{M-15-2.north east}{M-15-2.south east}{M-15-2.south west}
\DrawCell{M-15-3.north west}{M-15-3.north east}{M-15-3.south east}{M-15-3.south west}
\DrawCell{M-15-4.north west}{M-15-4.north east}{M-15-4.south east}{M-15-4.south west}
\DrawCell{M-15-5.north west}{M-15-5.north east}{M-15-5.south east}{M-15-5.south west}
\DrawCell{M-15-6.north west}{M-15-6.north east}{M-15-6.south east}{M-15-6.south west}

\node[anchor = north west, align=center, font=\footnotesize] at (M-2-2.north west) {\\ \textbf{\VarAA \%}};
\node[anchor = north west, align=center, font=\footnotesize] at (M-2-3.north west) {\\ \textbf{\VarAB \%}};
\node[anchor = north west, align=center, font=\footnotesize] at (M-2-4.north west) {\\ \textbf{\VarAC \%}};
\node[anchor = north west, align=center, font=\footnotesize] at (M-2-5.north west) {\\ \textbf{\VarAD \%}};
\node[anchor = north west, align=center, font=\footnotesize] at (M-2-6.north west) {\\ \textbf{ \VarAE \%}};

\node[anchor = north west, align=center, font=\footnotesize] at (M-3-2.north west) {\\ \textbf{ \VarBA \%}};
\node[anchor = north west, align=center, font=\footnotesize] at (M-3-3.north west) {\\ \textbf{ \VarBB \%}};
\node[anchor = north west, align=center, font=\footnotesize] at (M-3-4.north west) {\\ \textbf{\VarBC \%}};
\node[anchor = north west, align=center, font=\footnotesize] at (M-3-5.north west) {\\ \textbf{\VarBD \%}};
\node[anchor = north west, align=center, font=\footnotesize] at (M-3-6.north west) {\\ \textbf{\VarBE \%}};

\node[anchor = north west, align=center, font=\footnotesize] at (M-4-2.north west) {\\ \textbf{ \VarCA \%}};
\node[anchor = north west, align=center, font=\footnotesize] at (M-4-3.north west) {\\ \textbf{\VarCB \%}};
\node[anchor = north west, align=center, font=\footnotesize] at (M-4-4.north west) {\\ \textbf{\VarCC \%}};
\node[anchor = north west, align=center, font=\footnotesize] at (M-4-5.north west) {\\ \textbf{\VarCD \%}};
\node[anchor = north west, align=center, font=\footnotesize] at (M-4-6.north west) {\\ \textbf{ \VarCE \%}};

\node[anchor = north west, align=center, font=\footnotesize] at (M-5-2.north west) {\\ \textbf{\VarDA \%}};
\node[anchor = north west, align=center, font=\footnotesize] at (M-5-3.north west) {\\ \textbf{\VarDB \%}};
\node[anchor = north west, align=center, font=\footnotesize] at (M-5-4.north west) {\\ \textbf{\VarDC \%}};
\node[anchor = north west, align=center, font=\footnotesize] at (M-5-5.north west) {\\ \textbf{\VarDD \%}};
\node[anchor = north west, align=center, font=\footnotesize] at (M-5-6.north west) {\\ \textbf{ \VarDE \%}};

\node[anchor = north west, align=center, font=\footnotesize] at (M-6-2.north west) {\\ \textbf{ \VarEA \%}};
\node[anchor = north west, align=center, font=\footnotesize] at (M-6-3.north west) {\\ \textbf{\VarEB \%}};
\node[anchor = north west, align=center, font=\footnotesize] at (M-6-4.north west) {\\ \textbf{ \VarEC \%}};
\node[anchor = north west, align=center, font=\footnotesize] at (M-6-5.north west) {\\ \textbf{\VarED \%}};
\node[anchor = north west, align=center, font=\footnotesize] at (M-6-6.north west) {\\ \textbf{\VarEE \%}};

\node[anchor = north west, align=center, font=\footnotesize] at (M-7-2.north west) {\\ \textbf{\VarFA \%}};
\node[anchor = north west, align=center, font=\footnotesize] at (M-7-3.north west) {\\ \textbf{\VarFB \%}};
\node[anchor = north west, align=center, font=\footnotesize] at (M-7-4.north west) {\\ \textbf{\VarFC \%}};
\node[anchor = north west, align=center, font=\footnotesize] at (M-7-5.north west) {\\ \textbf{\VarFD \%}};
\node[anchor = north west, align=center, font=\footnotesize] at (M-7-6.north west) {\\ \textbf{ \VarFE \%}};

\node[anchor = north west, align=center, font=\footnotesize] at (M-8-2.north west) {\\ \textbf{ \VarGA \%}};
\node[anchor = north west, align=center, font=\footnotesize] at (M-8-3.north west) {\\ \textbf{ \VarGB \%}};
\node[anchor = north west, align=center, font=\footnotesize] at (M-8-4.north west) {\\ \textbf{\VarGC \%}};
\node[anchor = north west, align=center, font=\footnotesize] at (M-8-5.north west) {\\ \textbf{\VarGD \%}};
\node[anchor = north west, align=center, font=\footnotesize] at (M-8-6.north west) {\\ \textbf{\VarGE \%}};

\node[anchor = north west, align=center, font=\footnotesize] at (M-9-2.north west) {\\ \textbf{\VarHA \%}};
\node[anchor = north west, align=center, font=\footnotesize] at (M-9-3.north west) {\\ \textbf{\VarHB \%}};
\node[anchor = north west, align=center, font=\footnotesize] at (M-9-4.north west) {\\ \textbf{\VarHC \%}};
\node[anchor = north west, align=center, font=\footnotesize] at (M-9-5.north west) {\\ \textbf{\VarHD \%}};
\node[anchor = north west, align=center, font=\footnotesize] at (M-9-6.north west) {\\ \textbf{\VarHE \%}};

\node[anchor = north west, align=center, font=\footnotesize] at (M-10-2.north west) {\\ \textbf{\VarIA \%}};
\node[anchor = north west, align=center, font=\footnotesize] at (M-10-3.north west) {\\ \textbf{\VarIB \%}};
\node[anchor = north west, align=center, font=\footnotesize] at (M-10-4.north west) {\\ \textbf{\VarIC \%}};
\node[anchor = north west, align=center, font=\footnotesize] at (M-10-5.north west) {\\ \textbf{\VarID \%}};
\node[anchor = north west, align=center, font=\footnotesize] at (M-10-6.north west) {\\ \textbf{ \VarIE \%}};

\node[anchor = north west, align=center, font=\footnotesize] at (M-11-2.north west) {\\ \textbf{ \VarJA \%}};
\node[anchor = north west, align=center, font=\footnotesize] at (M-11-3.north west) {\\ \textbf{\VarJB \%}};
\node[anchor = north west, align=center, font=\footnotesize] at (M-11-4.north west) {\\ \textbf{\VarJC \%}};
\node[anchor = north west, align=center, font=\footnotesize] at (M-11-5.north west) {\\ \textbf{\VarJD \%}};
\node[anchor = north west, align=center, font=\footnotesize] at (M-11-6.north west) {\\ \textbf{ \VarJE \%}};

\node[anchor = north west, align=center, font=\footnotesize] at (M-12-2.north west) {\\ \textbf{\VarKA \%}};
\node[anchor = north west, align=center, font=\footnotesize] at (M-12-3.north west) {\\ \textbf{\VarKB \%}};
\node[anchor = north west, align=center, font=\footnotesize] at (M-12-4.north west) {\\ \textbf{\VarKC \%}};
\node[anchor = north west, align=center, font=\footnotesize] at (M-12-5.north west) {\\ \textbf{\VarKD \%}};
\node[anchor = north west, align=center, font=\footnotesize] at (M-12-6.north west) {\\ \textbf{\VarKE \%}};

\node[anchor = north west, align=center, font=\footnotesize] at (M-13-2.north west) {\\ \textbf{ \VarLA \%}};
\node[anchor = north west, align=center, font=\footnotesize] at (M-13-3.north west) {\\ \textbf{ \VarLB \%}};
\node[anchor = north west, align=center, font=\footnotesize] at (M-13-4.north west) {\\ \textbf{\VarLC \%}};
\node[anchor = north west, align=center, font=\footnotesize] at (M-13-5.north west) {\\ \textbf{\VarLD \%}};
\node[anchor = north west, align=center, font=\footnotesize] at (M-13-6.north west) {\\ \textbf{\VarLE \%}};

\node[anchor = north west, align=center, font=\footnotesize] at (M-14-2.north west) {\\ \textbf{ \VarMA \%}};
\node[anchor = north west, align=center, font=\footnotesize] at (M-14-3.north west) {\\ \textbf{\VarMB \%}};
\node[anchor = north west, align=center, font=\footnotesize] at (M-14-4.north west) {\\ \textbf{\VarMC \%}};
\node[anchor = north west, align=center, font=\footnotesize] at (M-14-5.north west) {\\ \textbf{\VarMD \%}};
\node[anchor = north west, align=center, font=\footnotesize] at (M-14-6.north west) {\\ \textbf{\VarME \%}};

\node[anchor = north west, align=center, font=\footnotesize] at (M-15-2.north west) {\\ \textbf{\VarNA \%}};
\node[anchor = north west, align=center, font=\footnotesize] at (M-15-3.north west) {\\ \textbf{\VarNB \%}};
\node[anchor = north west, align=center, font=\footnotesize] at (M-15-4.north west) {\\ \textbf{\VarNC \%}};
\node[anchor = north west, align=center, font=\footnotesize] at (M-15-5.north west) {\\ \textbf{\VarND \%}};
\node[anchor = north west, align=center, font=\footnotesize] at (M-15-6.north west) {\\ \textbf{\VarNE \%}};

\draw[opacity=\OPACITYODD,fill=OddColor] ($ (M-2-2.north west)!1 - \VarAA / 100!(M-2-2.south west) $) rectangle(M-2-2.south east);
\draw[opacity=\OPACITYODD,fill=OddColor] ($ (M-2-3.north west)!1 - \VarAB / 100!(M-2-3.south west) $) rectangle(M-2-3.south east);
\draw[opacity=\OPACITYODD,fill=OddColor] ($ (M-2-4.north west)!1 - \VarAC / 100!(M-2-4.south west) $) rectangle(M-2-4.south east);
\draw[opacity=\OPACITYODD,fill=OddColor] ($ (M-2-5.north west)!1 - \VarAD / 100!(M-2-5.south west) $) rectangle(M-2-5.south east);
\draw[opacity=\OPACITYODD,fill=OddColor] ($ (M-2-6.north west)!1 - \VarAE / 100!(M-2-6.south west) $) rectangle(M-2-6.south east);

\draw[opacity=\OPACITYEVEN,fill=EvenColor] ($ (M-3-2.north west)!1 - \VarBA / 100!(M-3-2.south west) $) rectangle(M-3-2.south east);
\draw[opacity=\OPACITYEVEN,fill=EvenColor] ($ (M-3-3.north west)!1 - \VarBB / 100!(M-3-3.south west) $) rectangle(M-3-3.south east);
\draw[opacity=\OPACITYEVEN,fill=EvenColor] ($ (M-3-4.north west)!1 - \VarBC / 100!(M-3-4.south west) $) rectangle(M-3-4.south east);
\draw[opacity=\OPACITYEVEN,fill=EvenColor] ($ (M-3-5.north west)!1 - \VarBD / 100!(M-3-5.south west) $) rectangle(M-3-5.south east);
\draw[opacity=\OPACITYEVEN,fill=EvenColor] ($ (M-3-6.north west)!1 - \VarBE / 100!(M-3-6.south west) $) rectangle(M-3-6.south east);

\draw[opacity=\OPACITYODD,fill=OddColor] ($ (M-4-2.north west)!1 - \VarCA / 100!(M-4-2.south west) $) rectangle(M-4-2.south east);
\draw[opacity=\OPACITYODD,fill=OddColor] ($ (M-4-3.north west)!1 - \VarCB / 100!(M-4-3.south west) $) rectangle(M-4-3.south east);
\draw[opacity=\OPACITYODD,fill=OddColor] ($ (M-4-4.north west)!1 - \VarCC / 100!(M-4-4.south west) $) rectangle(M-4-4.south east);
\draw[opacity=\OPACITYODD,fill=OddColor] ($ (M-4-5.north west)!1 - \VarCD / 100!(M-4-5.south west) $) rectangle(M-4-5.south east);
\draw[opacity=\OPACITYODD,fill=OddColor] ($ (M-4-6.north west)!1 - \VarCE / 100!(M-4-6.south west) $) rectangle(M-4-6.south east);

\draw[opacity=\OPACITYEVEN,fill=EvenColor] ($ (M-5-2.north west)!1 - \VarDA / 100!(M-5-2.south west) $) rectangle(M-5-2.south east);
\draw[opacity=\OPACITYEVEN,fill=EvenColor] ($ (M-5-3.north west)!1 - \VarDB / 100!(M-5-3.south west) $) rectangle(M-5-3.south east);
\draw[opacity=\OPACITYEVEN,fill=EvenColor] ($ (M-5-4.north west)!1 - \VarDC / 100!(M-5-4.south west) $) rectangle(M-5-4.south east);
\draw[opacity=\OPACITYEVEN,fill=EvenColor] ($ (M-5-5.north west)!1 - \VarDD / 100!(M-5-5.south west) $) rectangle(M-5-5.south east);
\draw[opacity=\OPACITYEVEN,fill=EvenColor] ($ (M-5-6.north west)!1 - \VarDE / 100!(M-5-6.south west) $) rectangle(M-5-6.south east);

\draw[opacity=\OPACITYODD,fill=OddColor] ($ (M-6-2.north west)!1 - \VarEA / 100!(M-6-2.south west) $) rectangle(M-6-2.south east);
\draw[opacity=\OPACITYODD,fill=OddColor] ($ (M-6-3.north west)!1 - \VarEB / 100!(M-6-3.south west) $) rectangle(M-6-3.south east);
\draw[opacity=\OPACITYODD,fill=OddColor] ($ (M-6-4.north west)!1 - \VarEC / 100!(M-6-4.south west) $) rectangle(M-6-4.south east);
\draw[opacity=\OPACITYODD,fill=OddColor] ($ (M-6-5.north west)!1 - \VarED / 100!(M-6-5.south west) $) rectangle(M-6-5.south east);
\draw[opacity=\OPACITYODD,fill=OddColor] ($ (M-6-6.north west)!1 - \VarEE / 100!(M-6-6.south west) $) rectangle(M-6-6.south east);

\draw[opacity=\OPACITYEVEN,fill=EvenColor] ($ (M-7-2.north west)!1 - \VarFA / 100!(M-7-2.south west) $) rectangle(M-7-2.south east);
\draw[opacity=\OPACITYEVEN,fill=EvenColor] ($ (M-7-3.north west)!1 - \VarFB / 100!(M-7-3.south west) $) rectangle(M-7-3.south east);
\draw[opacity=\OPACITYEVEN,fill=EvenColor] ($ (M-7-4.north west)!1 - \VarFC / 100!(M-7-4.south west) $) rectangle(M-7-4.south east);
\draw[opacity=\OPACITYEVEN,fill=EvenColor] ($ (M-7-5.north west)!1 - \VarFD / 100!(M-7-5.south west) $) rectangle(M-7-5.south east);
\draw[opacity=\OPACITYEVEN,fill=EvenColor] ($ (M-7-6.north west)!1 - \VarFE / 100!(M-7-6.south west) $) rectangle(M-7-6.south east);

\draw[opacity=\OPACITYODD,fill=OddColor] ($ (M-8-2.north west)!1 - \VarGA / 100!(M-8-2.south west) $) rectangle(M-8-2.south east);
\draw[opacity=\OPACITYODD,fill=OddColor] ($ (M-8-3.north west)!1 - \VarGB / 100!(M-8-3.south west) $) rectangle(M-8-3.south east);
\draw[opacity=\OPACITYODD,fill=OddColor] ($ (M-8-4.north west)!1 - \VarGC / 100!(M-8-4.south west) $) rectangle(M-8-4.south east);
\draw[opacity=\OPACITYODD,fill=OddColor] ($ (M-8-5.north west)!1 - \VarGD / 100!(M-8-5.south west) $) rectangle(M-8-5.south east);
\draw[opacity=\OPACITYODD,fill=OddColor] ($ (M-8-6.north west)!1 - \VarGE / 100!(M-8-6.south west) $) rectangle(M-8-6.south east);

\draw[opacity=\OPACITYEVEN,fill=EvenColor] ($ (M-9-2.north west)!1 - \VarHA / 100!(M-9-2.south west) $) rectangle(M-9-2.south east);
\draw[opacity=\OPACITYEVEN,fill=EvenColor] ($ (M-9-3.north west)!1 - \VarHB / 100!(M-9-3.south west) $) rectangle(M-9-3.south east);
\draw[opacity=\OPACITYEVEN,fill=EvenColor] ($ (M-9-4.north west)!1 - \VarHC / 100!(M-9-4.south west) $) rectangle(M-9-4.south east);
\draw[opacity=\OPACITYEVEN,fill=EvenColor] ($ (M-9-5.north west)!1 - \VarHD / 100!(M-9-5.south west) $) rectangle(M-9-5.south east);
\draw[opacity=\OPACITYEVEN,fill=EvenColor] ($ (M-9-6.north west)!1 - \VarHE / 100!(M-9-6.south west) $) rectangle(M-9-6.south east);

\draw[opacity=\OPACITYODD,fill=OddColor] ($ (M-10-2.north west)!1 - \VarIA / 100!(M-10-2.south west) $) rectangle(M-10-2.south east);
\draw[opacity=\OPACITYODD,fill=OddColor] ($ (M-10-3.north west)!1 - \VarIB / 100!(M-10-3.south west) $) rectangle(M-10-3.south east);
\draw[opacity=\OPACITYODD,fill=OddColor] ($ (M-10-4.north west)!1 - \VarIC / 100!(M-10-4.south west) $) rectangle(M-10-4.south east);
\draw[opacity=\OPACITYODD,fill=OddColor] ($ (M-10-5.north west)!1 - \VarID / 100!(M-10-5.south west) $) rectangle(M-10-5.south east);
\draw[opacity=\OPACITYODD,fill=OddColor] ($ (M-10-6.north west)!1 - \VarIE / 100!(M-10-6.south west) $) rectangle(M-10-6.south east);

\draw[opacity=\OPACITYEVEN,fill=EvenColor] ($ (M-11-2.north west)!1 - \VarJA / 100!(M-11-2.south west) $) rectangle(M-11-2.south east);
\draw[opacity=\OPACITYEVEN,fill=EvenColor] ($ (M-11-3.north west)!1 - \VarJB / 100!(M-11-3.south west) $) rectangle(M-11-3.south east);
\draw[opacity=\OPACITYEVEN,fill=EvenColor] ($ (M-11-4.north west)!1 - \VarJC / 100!(M-11-4.south west) $) rectangle(M-11-4.south east);
\draw[opacity=\OPACITYEVEN,fill=EvenColor] ($ (M-11-5.north west)!1 - \VarJD / 100!(M-11-5.south west) $) rectangle(M-11-5.south east);
\draw[opacity=\OPACITYEVEN,fill=EvenColor] ($ (M-11-6.north west)!1 - \VarJE / 100!(M-11-6.south west) $) rectangle(M-11-6.south east);

\draw[opacity=\OPACITYODD,fill=OddColor] ($ (M-12-2.north west)!1 - \VarKA / 100!(M-12-2.south west) $) rectangle(M-12-2.south east);
\draw[opacity=\OPACITYODD,fill=OddColor] ($ (M-12-3.north west)!1 - \VarKB / 100!(M-12-3.south west) $) rectangle(M-12-3.south east);
\draw[opacity=\OPACITYODD,fill=OddColor] ($ (M-12-4.north west)!1 - \VarKC / 100!(M-12-4.south west) $) rectangle(M-12-4.south east);
\draw[opacity=\OPACITYODD,fill=OddColor] ($ (M-12-5.north west)!1 - \VarKD / 100!(M-12-5.south west) $) rectangle(M-12-5.south east);
\draw[opacity=\OPACITYODD,fill=OddColor] ($ (M-12-6.north west)!1 - \VarKE / 100!(M-12-6.south west) $) rectangle(M-12-6.south east);

\draw[opacity=\OPACITYEVEN,fill=EvenColor] ($ (M-13-2.north west)!1 - \VarLA / 100!(M-13-2.south west) $) rectangle(M-13-2.south east);
\draw[opacity=\OPACITYEVEN,fill=EvenColor] ($ (M-13-3.north west)!1 - \VarLB / 100!(M-13-3.south west) $) rectangle(M-13-3.south east);
\draw[opacity=\OPACITYEVEN,fill=EvenColor] ($ (M-13-4.north west)!1 - \VarLC / 100!(M-13-4.south west) $) rectangle(M-13-4.south east);
\draw[opacity=\OPACITYEVEN,fill=EvenColor] ($ (M-13-5.north west)!1 - \VarLD / 100!(M-13-5.south west) $) rectangle(M-13-5.south east);
\draw[opacity=\OPACITYEVEN,fill=EvenColor] ($ (M-13-6.north west)!1 - \VarLE / 100!(M-13-6.south west) $) rectangle(M-13-6.south east);

\draw[opacity=\OPACITYODD,fill=OddColor] ($ (M-14-2.north west)!1 - \VarMA / 100!(M-14-2.south west) $) rectangle(M-14-2.south east);
\draw[opacity=\OPACITYODD,fill=OddColor] ($ (M-14-3.north west)!1 - \VarMB / 100!(M-14-3.south west) $) rectangle(M-14-3.south east);
\draw[opacity=\OPACITYODD,fill=OddColor] ($ (M-14-4.north west)!1 - \VarMC / 100!(M-14-4.south west) $) rectangle(M-14-4.south east);
\draw[opacity=\OPACITYODD,fill=OddColor] ($ (M-14-5.north west)!1 - \VarMD / 100!(M-14-5.south west) $) rectangle(M-14-5.south east);
\draw[opacity=\OPACITYODD,fill=OddColor] ($ (M-14-6.north west)!1 - \VarME / 100!(M-14-6.south west) $) rectangle(M-14-6.south east);

\draw[opacity=\OPACITYEVEN,fill=EvenColor] ($ (M-15-2.north west)!1 - \VarNA / 100!(M-15-2.south west) $) rectangle(M-15-2.south east);
\draw[opacity=\OPACITYEVEN,fill=EvenColor] ($ (M-15-3.north west)!1 - \VarNB / 100!(M-15-3.south west) $) rectangle(M-15-3.south east);
\draw[opacity=\OPACITYEVEN,fill=EvenColor] ($ (M-15-4.north west)!1 - \VarNC / 100!(M-15-4.south west) $) rectangle(M-15-4.south east);
\draw[opacity=\OPACITYEVEN,fill=EvenColor] ($ (M-15-5.north west)!1 - \VarND / 100!(M-15-5.south west) $) rectangle(M-15-5.south east);
\draw[opacity=\OPACITYEVEN,fill=EvenColor] ($ (M-15-6.north west)!1 - \VarNE / 100!(M-15-6.south west) $) rectangle(M-15-6.south east);
\end{tikzpicture}
\footnotesize{A response of 1 corresponds to strongly disagree whilst a response of 5 corresponds to strongly agree.}
\end{figure}

Out of those who left additional comments, 50\% made a remark about the overall difficulty of the task generally pertaining to the lack of depth perception. 
21\% of those who left comments also mentioned being “not sure what the exact behaviour of the assistance is”. Participants were not provided with information about the methodology of the assistant, causing some participants to believe that the assistant dampened movement whilst landing for finer control or prevented control inputs from being registered when landing where no platform exists. 

A further 17\% of those who left comments made remarks regarding the inconsistency of the assistant in terms of the degree of assistance provided. It was observed that the assistant would automatically land the drone for the pilot if the assistant was confident about the pilot's goal. However if the assistant was not confident in its estimate of the pilot's goal, the assistant would instead adjust the pilot's input trajectory.

\begin{figure}
\centering
\includegraphics[width=1.0\columnwidth]{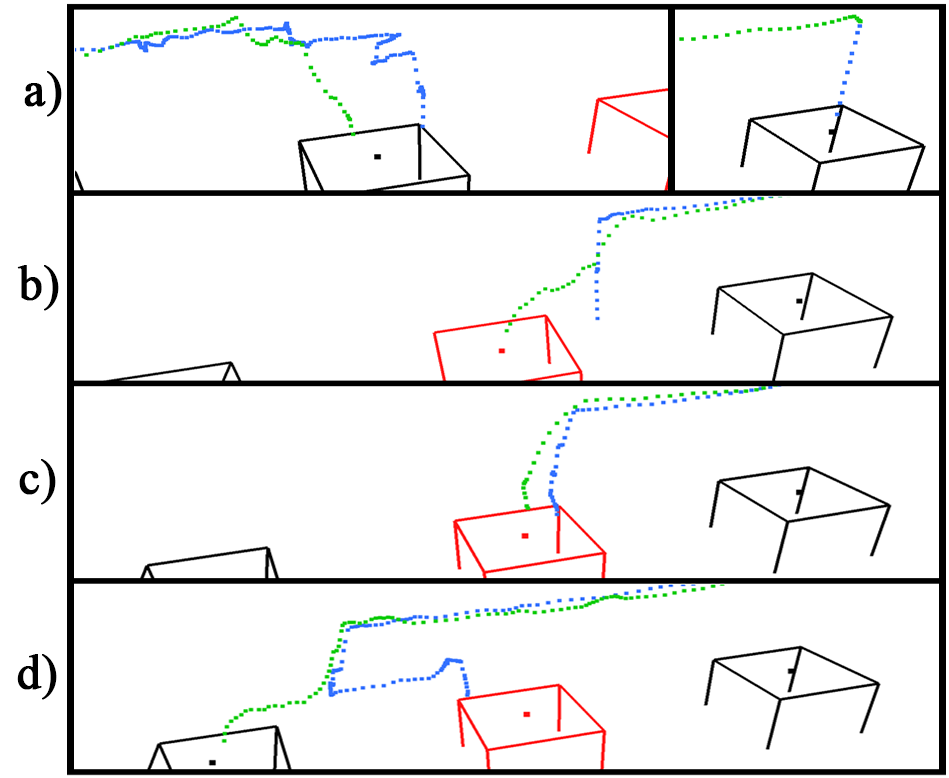}
\caption{Trajectories of various participants landing whilst unassisted (blue) and assisted (green) where the goal platform is denoted in red.  Examples a), b) \& c) show an inexperienced, intermediate and expert user respectively performing a landing. Example d) demonstrates a failure case for the assistant.}
\label{LandingExamples}
\end{figure}

Fig.~\ref{LandingExamples} demonstrates how various skill level participants land the drone. Inexperienced users as shown in Fig.~\ref{LandingExamples} a) tended to have intermittent movement from briefly tapping the movement keys whilst frequently landing near incorrect platforms. Inexperienced participants were also observed to get confused with the control scheme as seen in the right of a) where the participant mistook the fly backwards key for the fly down key. Intermediate participants as shown in b) were able to land in the general vicinity of the landing platform but required assistance to safely land. Experienced participants c) reliably landed close to the platforms and only required the assistant for fine movement control. Example trajectory d) shows a failure case of the assistant for an intermediate participant where the participant attempted to land in the centre of the valley between two platforms, leading to the assistant incorrectly inferring the participant’s true intent and prioritised preventing a crash over conforming to the pilots actions. 

Incorrectly inferring the pilot’s intent may lead to unsatisfactory results however in applied scenarios the pilot will favour a safe landing on an incorrect location over a crash. Whilst assisted, participants had a crash rate of 6.24\% compared to 70.50\% unassisted. It was observed by analysing the previous 3 seconds prior to the assistant landing on the incorrect platform that 67.62\% of participants were attempting to descend. Due to the challenging depth perception task, novice participants may not have been aware that they were heading towards an incorrect target, limiting the time for the assistant to reconfirm the pilot’s intent whilst prioritising a safe landing. A further 20.95\% of participants performed little or no actions whilst the assistant guided the UAV to the incorrect platform, resulting in insufficient evidence for the assistant to infer their true intent.

\begin{figure}[!t]
\centering
\includegraphics[width=1.0\columnwidth]{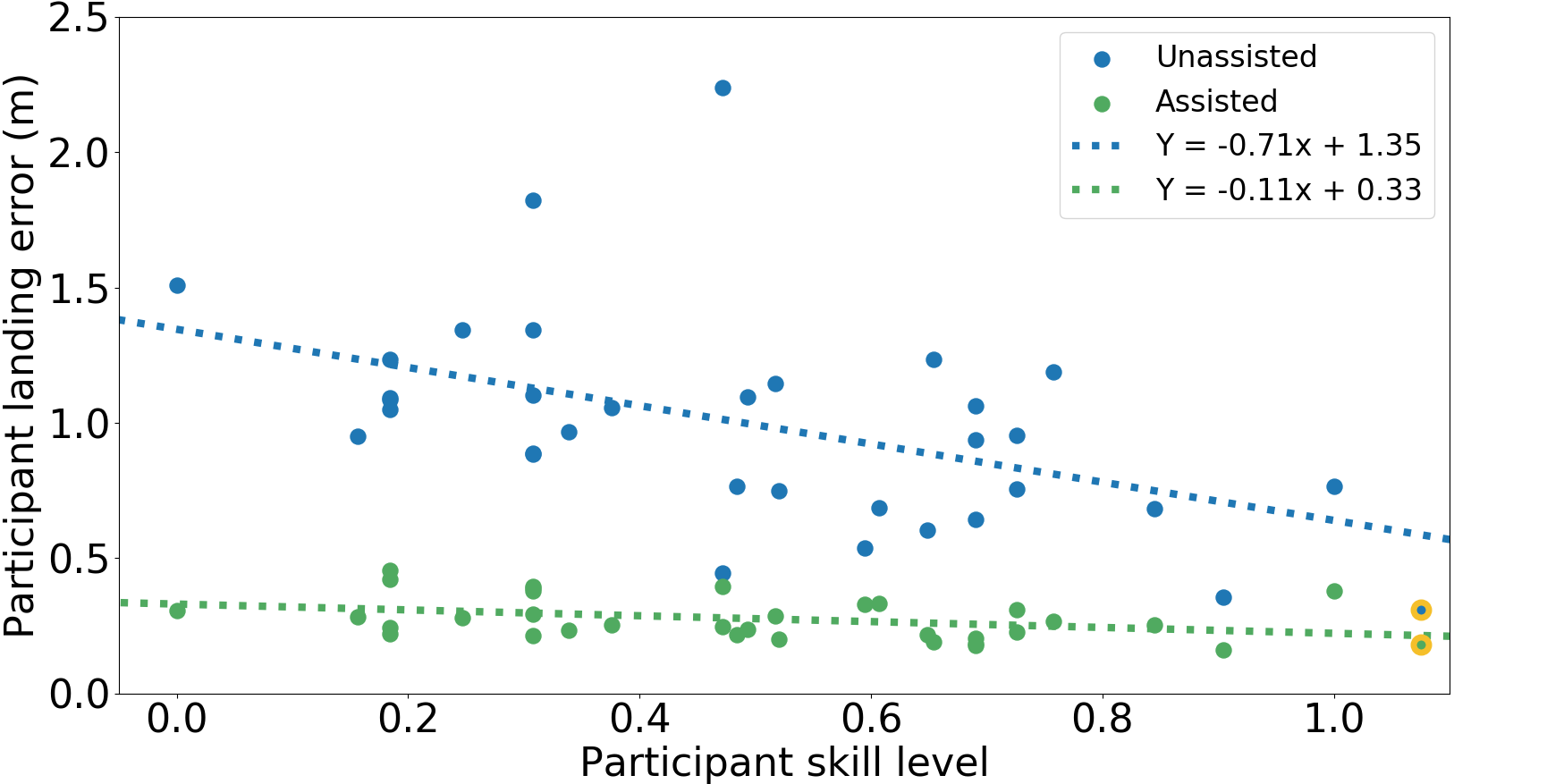}
\caption{Plot of participants' median error for flying unassisted (blue) and assisted (green) against their corresponding skill level. The first authors attempt is shown highlighted in yellow and is excluded from all forms of statistical analysis.  }
\label{UserStudyRegression}
\end{figure}

To quantify the level of assistance provided and to verify if the assistant allows novice pilots to perform at a level equal to or greater then experienced pilots, we plot each participant's median error against their experience level as shown in Fig.~\ref{UserStudyRegression}. To determine a pilot’s experience level, we implement a linear regression model that fits the pilot's average landing error as a dependant variable whilst using responses to questions 2, 3, 5 \& 6 in Table.~\ref{UserStudyDemoTable} as independent variables. The regressed model provides an estimate of what the predicted mean error should be for a pilot of given experience. The predicted error is normalised between 0 and 1, then inverted so that an experience level of 0 represents a novice whilst an experience level of 1 represents an expert. Fig.~\ref{UserStudyRegression} shows that whilst assisted, regardless of one’s skill level, participants will perform at an equal proficiency that is greater than skilled pilots. Whilst assisted, participant performance was not greatly dependant on their individual skill level which is reflected by a low magnitude regression coefficient of -0.11 compared to that of -0.71 whilst being unassisted. The first author's attempt at the user study is included in Fig.~\ref{UserStudyRegression}, indicated by the yellow highlighted point. This data point is excluded from all statistical analysis but is shown to demonstrate the potential performance of a pilot who has had substantial time to become familiar with the specific setup. 

\section{Physical demonstration}
\begin{figure}[!t]
\centering
\includegraphics[width=1.0\columnwidth]{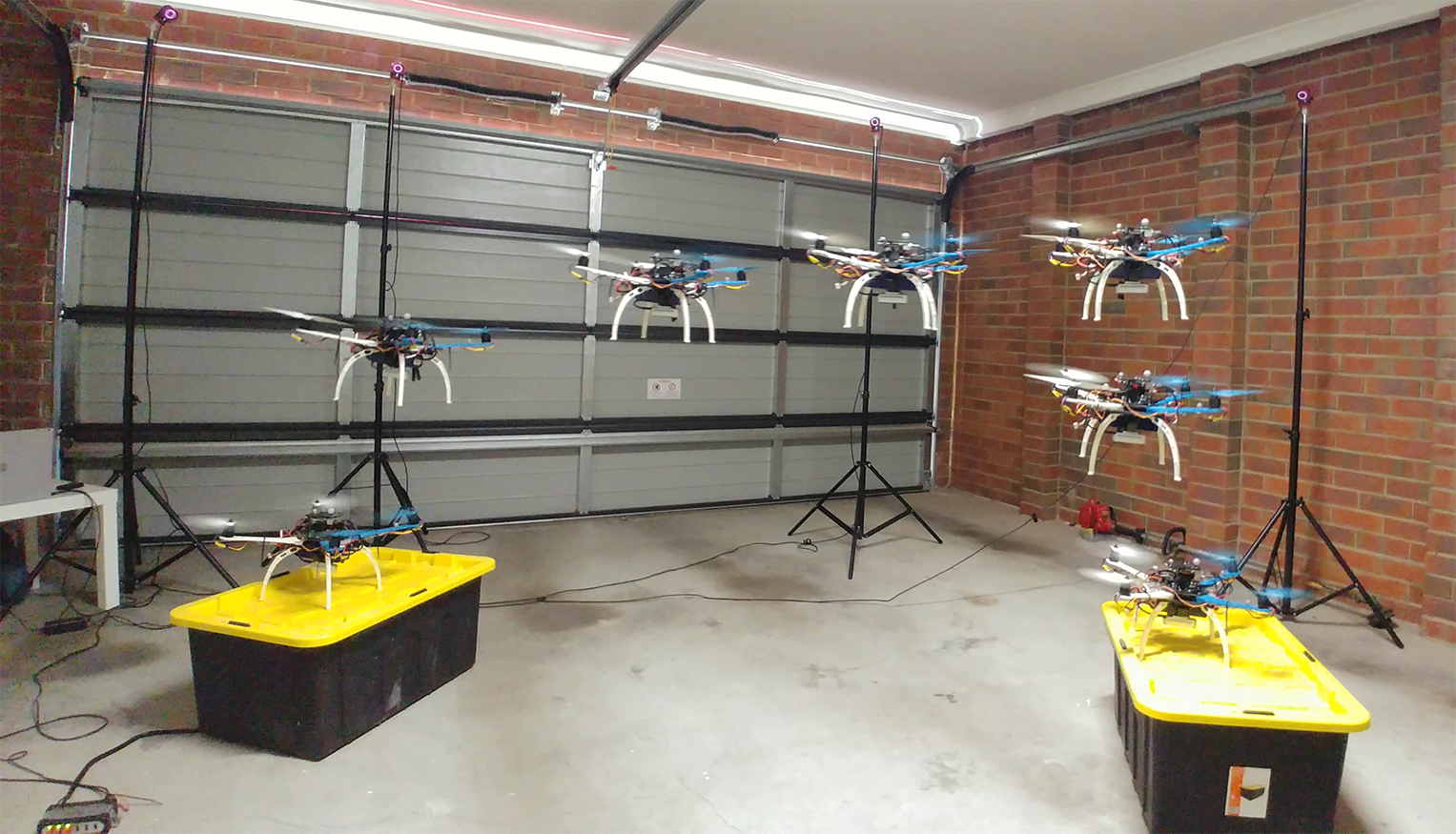}
\caption{Trajectory of pilot landing whilst assisted. The pilot initially undershoots the platform before being corrected by the assistant.}
\label{PhysicalDrone}
\end{figure}

To show the feasibility of the assistant, we integrated the approach onto a custom-built UAV using the Pixhawk 2 flight controller running ArduPilot. The drone pose was estimated using an OptiTrack motion capture system. A Raspberry Pi 3b+ onboard companion computer was used for wireless communication and sending RGB-D images from an Intel RealSense D435i camera to a base station for network inference. The environment consisted of two platforms due to spatial constraints. The pilot was tasked with landing on the furthest platform whilst unassisted and assisted. The policy network was retrained using a CM-VAE that was trained on a collection of synthetic and real images that match the intrinsic parameters of the camera used. Aside from the inclusion of additional training data, the network structure and parameters were identical to the simulation study. The final output velocity of the assistant and pilot was halved due to safety concerns. The assistant was able to guide the UAV towards the centre of the platform as seen in Fig.~\ref{PhysicalDrone} without the need for modifications to the proposed approach in the hosted user study. The assisted condition resulted in a smoother trajectory due to the continuous action space provided by the assistant compared to the discrete key presses available to the pilot. A limitation of the assistant was with handling the interaction between the UAV legs and the platform, as this was not modelled in simulation. The supplementary video shows an example of the physical demonstration.

\section{Conclusion}
In this work we present a shared autonomy approach that allows novice pilots to land a drone on one of several platforms at a proficiency greater than experienced pilots. The approach contains two components (i) an encoder that compresses RGB-D information into a latent vector to facilitate the perception of the physical environment and (ii) a policy network that provides control inputs to assist the pilot with landing. The encoder is trained using a cross-modal variational auto-encoder which takes noisy RGB-D data and reconstructs a denoised depth map. The policy network is trained within simulation using the reinforcement learning algorithm TD3.  
In a user study (n=33), participants tasked with landing a simulated drone on a specified platform achieved significantly higher success rates and lower median and mean errors in the assisted condition, compared to unassisted. Participants also preferred the assistant for performing the task and perceived it to be more usable. However the success of landing on the correct platform is limited by the proficiency of the assistant at estimating the true intent of the pilot. Regardless of skill level, participants performed better assisted than an unassisted experienced participant, confirming that this approach allows novice pilots to perform at a level greater than or equal to that of experienced pilots. An initial physical demonstrator illustrates that the proposed approach can be deployed on a physical platform.

Future work is planned to integrate assisted UAV landing to physical environments with extensive user study validation whilst addressing the consistency concerns brought up by participants.

\bibliographystyle{./bibliography/IEEEtran}
\bibliography{./bibliography/IEEEexample}

\end{document}